\documentclass[journal]{IEEEtran}

\usepackage{caption}
\usepackage{subcaption}
\ifCLASSINFOpdf
  \usepackage[pdftex]{graphicx}
  \graphicspath{{./graph/}}
  \DeclareGraphicsExtensions{.pdf,.jpeg,.png}
\else
  \graphicspath{{./graph/}}
  \DeclareGraphicsExtensions{.eps}
\fi

\ifCLASSINFOpdf
\else
\fi

\usepackage[cmex10]{amsmath,mathtools}
\usepackage{mdwmath}
\usepackage{mdwtab}
\usepackage{amssymb}
\usepackage{bm}

\usepackage{algorithmic}
\usepackage[ ruled, vlined]{algorithm2e}
\usepackage{float}

\usepackage{epstopdf}
\usepackage{color}

\usepackage{cite}

\usepackage{makecell}

\begin{document}

\title{Exploring Deep Reinforcement Learning-Assisted Federated Learning for Online Resource Allocation in Privacy-Preserving EdgeIoT}

\author{Jingjing Zheng, Kai Li,~\IEEEmembership{Senior Member,~IEEE,} Naram Mhaisen, Wei Ni,~\IEEEmembership{Senior Member,~IEEE,} Eduardo Tovar, and Mohsen Guizani,~\IEEEmembership{Fellow,~IEEE,}
\thanks{J.~Zheng, K.~Li and E.~Tovar are with the Real-Time and Embedded Computing Systems Research Centre (CISTER), 4249--015 Porto, Portugal (E-mail: \{zheng,kai,emt\}@isep.ipp.pt).}

\thanks{Naram Mhaisen is with the College of Electrical Engineering, Mathematics, and Computer Science (EEMCS), TU Delft, Netherlands (e-mail: n.mhaisen@tudelft.nl).}

\thanks{Wei Ni is with the Commonwealth Scientific and Industrial Research Organisation (CSIRO), Sydney 2122, Australia (E-mail: wei.ni@data61.csiro.au).}

\thanks{ Mohsen Guizani is with the Machine Learning Department, Mohamed Bin Zayed University of Artificial Intelligence (MBZUAI), Abu Dhabi, UAE (E-mail: mguizani@ieee.org).}

\thanks{Corresponding author: Kai Li.}
}

\markboth{IEEE INTERNET OF THINGS JOURNAL,~2022.}%
{Jingjing \MakeLowercase{\textit{et al.}}:Exploring deep reinforcement learning-assisted federated learning for online resource allocation in Privacy-Preserving EdgeIoT}

\IEEEcompsoctitleabstractindextext{%
\begin{abstract}
\boldmath
Federated learning (FL) has been increasingly considered to  preserve data training privacy from eavesdropping attacks in
mobile edge computing-based Internet of Thing (EdgeIoT). On the one hand, the learning accuracy of FL can be improved by selecting the IoT devices with large datasets for training, which gives rise to a higher energy consumption. On the other hand, the energy consumption can be reduced by selecting the IoT devices with small datasets for FL, resulting  in a falling learning accuracy. In this paper, we formulate a new resource allocation problem for privacy-preserving EdgeIoT to balance the learning accuracy of FL and the energy consumption of the IoT device. 
We propose a new federated learning-enabled twin-delayed deep deterministic policy gradient (FL-DLT3) framework to achieve the optimal accuracy and energy balance in a continuous domain. Furthermore, long short term memory (LSTM) is leveraged in FL-DLT3 to predict the time-varying network state while FL-DLT3 is trained to select the IoT devices and allocate the transmit power. Numerical results demonstrate that the proposed FL-DLT3 achieves fast convergence (less than 100 iterations) while the FL accuracy-to-energy consumption ratio is improved by 51.8\% compared to existing state-of-the-art benchmark. 
\end{abstract}

}

\maketitle
\IEEEdisplaynotcompsoctitleabstractindextext
\IEEEpeerreviewmaketitle    
\section{Introduction} \label{sec_introduction}
 Mobile  edge computing (MEC) provides a promising solution to enabling cloud computing services in privacy-persevering mobile edge computing-based Internet of Thing (EdgeIoT) \cite{mao2017survey, sun2016edgeiot, Deng9001216}. The IoT devices can offload their local computation-intensive tasks to computationally powerful edge servers \cite{mach2017mobile, 9352033}. In EdgeIoT, offloading the source data of the IoT devices to the edge server is vulnerable to eavesdropping attacks \cite{Liangxiao,9183799}. To prevent private data leakage of the IoT devices, federated learning (FL) \cite{mcmahan2017communication} is used to train a global data learning model at the edge server by aggregating the data structure parameters of the IoT devices, while the source data remains at the IoT devices.  

\begin{figure}[htbp!]
\centerline{\includegraphics[width = 0.5\textwidth]{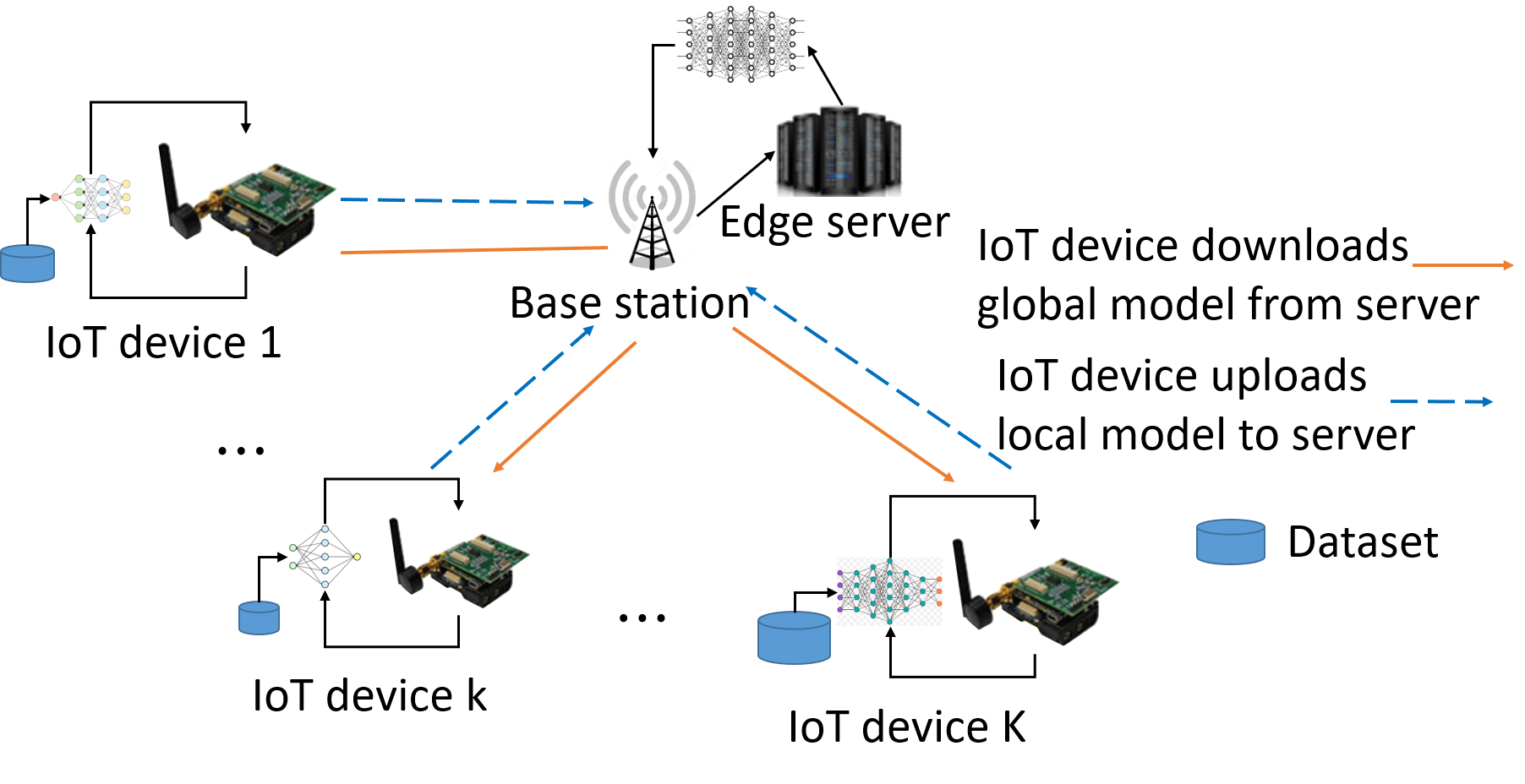}}
 \caption{FL-enabled EdgeIoT. A local model is trained with the dataset at the IoT device.  The local model is aggregated by an edge server, where the global model is trained and returned to the IoT devices.}
 \label{fig:1-1}
 \end{figure}

Fig. \ref{fig:1-1} depicts an FL-enabled EdgeIoT, where the IoT devices are deployed to sense and process private information,  e.g., health reports of patients \cite{xiao2018iot}.  Specifically, mobile phones use image classification models to classify pictures or images. FL can help multiple mobile phones to cooperatively train an effective global image classification model, without the need of sharing the images and pictures used for the training \cite{Ji2019}. Likewise, vehicles can train an accurate autonomous driving model by  aggregating the local models trained separately by multiple vehicles  \cite{LiYijing2021}.

 A local  model (e.g., the weight vector \cite{McMahanMRA16, kang2019, Lim2021}  or gradient \cite{ji2019learning}) is trained on the sensing  data of an IoT device to sense and process private information, such as private health reports of patients. Next, the edge server aggregates the local models of all IoT devices to create a comprehensive and effective global model without collecting the private data of the IoT devices.  A global model is obtained at the edge server, e.g., by applying FedAvg \cite{McMahanMRA16} to  the local models. The edge server broadcasts the  global model back to all the IoT devices. According to the global model, each of the IoT devices renews the training of its local model. By iteratively training the local model at the IoT device and updating the global model at the edge server, the learning accuracy of FL on the data classification and event prediction can be progressively improved \cite{9220780}.

 While selecting the IoT devices with large training datasets can improve the learning accuracy of FL \cite{Dinesh}, it can often result in fast depletion of the batteries at the devices. On the other hand, selecting the IoT devices with small data for training the local  models  can save the battery energy of the IoT devices, but likely leads to a low accuracy of the global model. In this sense,  balancing the learning accuracy of FL and energy consumption of the IoT devices is crucial to EdgeIoT. 

 Moreover, the FL process is time-slotted by design and  sequential decision making is required. Hence, in this paper, we propose an online resource allocation optimization to balance the learning accuracy of FL and energy consumption of the IoT devices. In practice, the instantaneous information of data size, transmit power, and link qualities between the edge server and the IoT devices is unlikely to be known.  The optimization is formulated as a partially observable Markov decision process (POMDP),  where the network state consists of the source data size of the IoT devices, channel conditions between the edge server and the IoT devices, bandwidth and remaining energy of the selected IoT devices. All the entries of a network state are continuous. The action space includes the discrete selection of IoT devices and the continuous transmit power allocation. Given the continuous network states and actions, a large number of IoT nodes lead to a large state and action space, and complex network state transitions. This results in difficulty in the joint optimization of IoT device selection and transmit  power allocation. Due to a large and continuous state and action space in the formulated POMDP, a new  deep reinforcement learning-based device selection and transmit power allocation algorithm is proposed to maximize the ratio of the learning accuracy of FL and  the energy consumption of the IoT devices. The major contributions of this article are summarized as follows:
\begin{itemize}
\item We propose to jointly optimize the selection of IoT devices and their transmit powers in an FL-empowered edge IoT system, thereby balancing  the learning accuracy of FL and energy consumption of the IoT devices. The optimization is online, adapting to the time-varying arrivals of training data and the energy budget of the IoT devices, and the changing channel conditions between the IoT devices and the base station (i.e., the model aggregator).

\item FL-DLT3 is proposed to learn the network state dynamics while maximizing the ratio of the learning accuracy of FL to the energy consumption of the IoT devices. Considering the continuous transmit powers of the IoT devices, FL-DLT3 optimizes the edge server's selection of IoT devices for each round of FL and the transmit powers of the IoT devices,  based on Twin Delayed Deep Deterministic Policy Gradient (TD3).

\item A new long short term memory (LSTM) layer is designed in coupling with the proposed FL-DLT3 to predict the time-varying network states, e.g., data size, bandwidth, channel gain, and remaining energy of the IoT devices. The LSTM layer estimates the unobserved states at every training iteration of the TD3. To the best of our knowledge, this is the first time that LSTM is employed in coupling  with TD3 for the resource allocation of FL-enabled EdgeIoT.

\item FL-DLT3 is implemented in PyTorch. The effectiveness of FL-DLT3 is validated with the experimental data. Numerical results show that FL-DLT3 achieves fast convergence (less than 100 iterations) while the FL-accuracy-to-energy-consumption ratio is improved by 51.8\%, as compared to the state of the art. 
\end{itemize}

The rest of this paper is structured as follows. The literature on FL-based resource allocation in MEC is reviewed in Section~\ref{sec_related-work}. Section~\ref{sec_FL-based}  presents the FL protocol and system models. The resource allocation optimization for EdgeIoT is formulated in Section~\ref{sec_problem_formu}. Section~\ref{sec_drl-based} proposes the FL-DLT3 framework which conducts deep reinforcement learning (DRL) based EdgeIoT devices selection and resource allocation. Section~\ref{section-numerical} evaluates the proposed FL-DLT3 framework. Finally, Section~\ref{sec_conclusion} concludes this paper.


\section{Related Work} \label{sec_related-work}
This section presents the literature on resource allocation with FL in MEC. 

Assuming that the IoT devices have the same computational resources and wireless channel conditions, Federated averaging (FedAvg) \cite{McMahanMRA16} randomly selects IoT devices to participate FL training and synchronously aggregates local models. This process is repeated until a desirable training accuracy is achieved. In \cite{Nishio2019},  different IoT devices owns different computing capabilities and wireless channel conditions, resulting in different local model training time and upload time. The authors develop a FL protocol called FedCS, which allows the server to aggregate as many local model updates as possible to improve image classification accuracy.

To improve the training accuracy of FL systems in the
context of wireless channels and energy arrivals of
mobile devices, the authors \cite{Shunfeng2021}  model the transmission power allocation and mobile device selection of the FL training as a constrained Markov decision process. Due to a high complexity, stochastic learning methods and Lagrange multipliers are used to simplify the model and  to obtain an efficient policy for all mobile device. In \cite{AnhLNKW19}, the devices with limited battery energy, CPU computations, and bandwidths, are considered in a mobile crowd network. A deep Q-learning-based resource allocation for data, energy, and CPU cycles is developed to reduce the energy consumption of FL-based mobile devices and training time of the FL.  Given limited computation and communication resources at the devices, \cite{wang2019adaptive} analyzes the convergence bound of distributed gradient descent. A control algorithm is developed to determine the tradeoff between local update and global parameter aggregation. Some IoT devices have limited communication and computing resources and fail to complete training tasks, which leads to many discarded learning rounds affecting the model accuracy \cite{Wahab2021}. The authors study multicriteria-based approach for IoT device selection in FL that reducing the number of communication rounds to reach the intended accuracy and increasing the number of selected  IoT devices in each round.

The authors of \cite{yoshida2020mab} study a trial-and-error based IoT device selection  based on  multi-armed bandit (MAB), where computation tasks, traffic, and link qualities are unknown. The MAB-based IoT device selection balances the IoT device selection according to the selection frequency and IoT devices' resources.  In \cite{xu2021online}, the IoT device scheduling in MEC is formulated without the IoT devices' channel and computing information. To reduce the training latency, the IoT device scheduling is formulated as a MAB problem, where $\epsilon$-greedy is used to reduce the learning accuracy. In \cite{xia2020}, the authors develop a MAB-based IoT device scheduling framework to reduce the training latency of FL. Given the known independent and identically distributed (i.i.d.) local data at the IoT devices, an FL-based IoT device scheduling algorithm is designed.

 The authors of \cite{zhu2021federated} aim to reduce the
average age of data sources by controlling the IoT devices,
scheduling the data and allocating the bandwidth. An actor-critic learning framework is developed, where the IoT devices learn the scheduling strategies based on their local observations. To reduce the  training time and energy consumption  of IoT devices, the authors of \cite{zhan9139873} design an experience-driven algorithm based on proximal policy optimization and produce sub-optimal results.  To improve the network throughput,  the authors of \cite{kwon9067847} present  a  cell association and base station allocation method. Multi-agent deep deterministic policy gradient  is used to handle unexpected events and unreliable channels in underwater wireless networks.  In \cite{MeixiaTao}, the IoT devices are selected to improve the FL accuracy and reduce the training time and energy consumption. To balance the training accuracy and delay  of FL and  energy consumption, Twin Delayed Deep Deterministic policy gradient algorithm (TD3) is employed to capture the interplay between function approximation error in both policy and value updates, and produce the  IoT devices scheduling policy, the CPU frequency allocated for training, and the transmit power allocation.

An FL-based device selection optimization is developed in our preliminary work \cite{zheng2021federated} to balance the energy consumption of the IoT devices and the
learning accuracy of FL. The optimization model takes
advantage of the a-priori knowledge of the network state
information, e.g., data size, bandwidth, and channel gain. Due to NP-hardness of the optimization, an energy efficiency-FL accuracy balancing heuristic algorithm (FedAECS) was presented in \cite{zheng2021federated} to approximate
the optimal IoT device selection policy offline. In contrast, this paper considers a practical scenario without the prior  information about data size, bandwidth, channel gain, and remaining energy of the IoT devices. Due to a large state and action space, we propose a new FL-DLT3 to balance FL accuracy and energy consumption of the IoT devices online, where LSTM is leveraged to predict the hidden state of the IoT devices as input state of TD3. In addition,  we also compare the performance of the proposed FL-DLT3 with the FedAECS in \cite{zheng2021federated}.

\section{System Model}  \label{sec_FL-based}
In this section, we study the training protocol and energy model of FL. The notations used in the paper are summarized in Table \ref{tab:table3-1}.
\begin{table}[h!]
\small 
\begin{center}
    \caption{The list of key variables defined in system model}
    \label{tab:table3-1}
    \begin{tabular}{|c|c|} 
     \hline
     \textbf{Notation} & \textbf{Definition} \\
      \hline
      $K$  & The total number of IoT devices \\
        \hline
        $k$ & Index of IoT device \\
            \hline
       $T$ &  The total number of the rounds\\
       \hline
        $t$ & Index of the rounds \\
       \hline
         $D_{t,k}$ &  IoT device $k$ owns data size \\
           \hline
       $\mathbf{x}_{ki}$ & Input of the FL model \\
         \hline
       $y_{ki}$ & Output of the FL model \\
         \hline
      $\mathbf{w}$ & Weight parameter of FL training\\
       \hline
    $ \beta_{t,k}$ &  Whether IoT device $k$ is selected\\
      \hline
      $f_k$ &   CPU frequency of IoT device $k$\\
       \hline
      $\zeta_k$  & Effective capacitance coefficient\\
     \hline
     $L$ & Number of local iterations of FL training  \\
       \hline
      $E_{t,k1}^{cmp}$ & \makecell[c]{ Energy consumption for computing \\ $c_{k}D_{t,k}$ CPU cycles} \\
      \hline
      $  E_{t,k}^{cmp} $ & \makecell[c]{Total computation energy at \\ IoT device $k$ }  \\
      \hline
       $r_{t,k}^{up}$ &  Achievable uplink transmit rate \\
       \hline
      $b_{t.k}$ & \makecell[c]{Bandwidth allocated to IoT \\ device $k$ by the server }\\
      \hline
         $P_{t,k}$ & IoT device $k$ transmit power \\
       \hline
     $G_{t,k}$ & Uplink channel gain \\
      \hline
    $r_{t,k}^{down}$ & \makecell[c]{Achievable downlink transmit rate \\ of IoT device $k$  }\\
    \hline
    $H_{t,k}$ &  Downlink channel gain \\
     \hline 
       $P_t^s$ &    Transmit power of the edge server \\
       \hline
    $\tau_{t,k}^{down}$ & \makecell[c]{Downloading time of the global model \\ at IoT device $k$ } \\
     \hline
       $\mathfrak{S}_{d}$ &   Size of the global model \\ 
        \hline
    $\tau_{t,k}^{up}$ & Transmission time of the local model at $t$ \\
    \hline 
       $\mathfrak{S}_{u}$ &   Size of the local model \\ 
     \hline
     $E_{t,k}^{up}$ & \makecell[c]{ Energy consumption of device $k$ \\ on the local model transmission }\\
     \hline
   $E_{t,k}^{c}$ &  \makecell[c]{Total energy consumption  of \\ the selected  device $k$ } \\
     \hline
   $E_{t,k}$ &  Remaining battery energy of device $k$ \\
     \hline
    $\Delta E_{t,k}$ &  Amount of harvested energy \\
    \hline
    $\Delta_t$ &  Data evenness of selected devices \\
    \hline
       $\tau_{t,k} $ & \makecell[c]{Completion time of IoT \\ device $k$ in $t$-th round}\\
       \hline
      $\mu_{k}$  & System parameter \\
      \hline
       $\nu$ & \makecell[c]{Constant value}\\
       \hline
    \end{tabular}
 \end{center}
\end{table}

\subsection{FL Protocol with MEC}
\begin{figure*}[htbp!]
\centerline{\includegraphics[width = 0.8\textwidth]{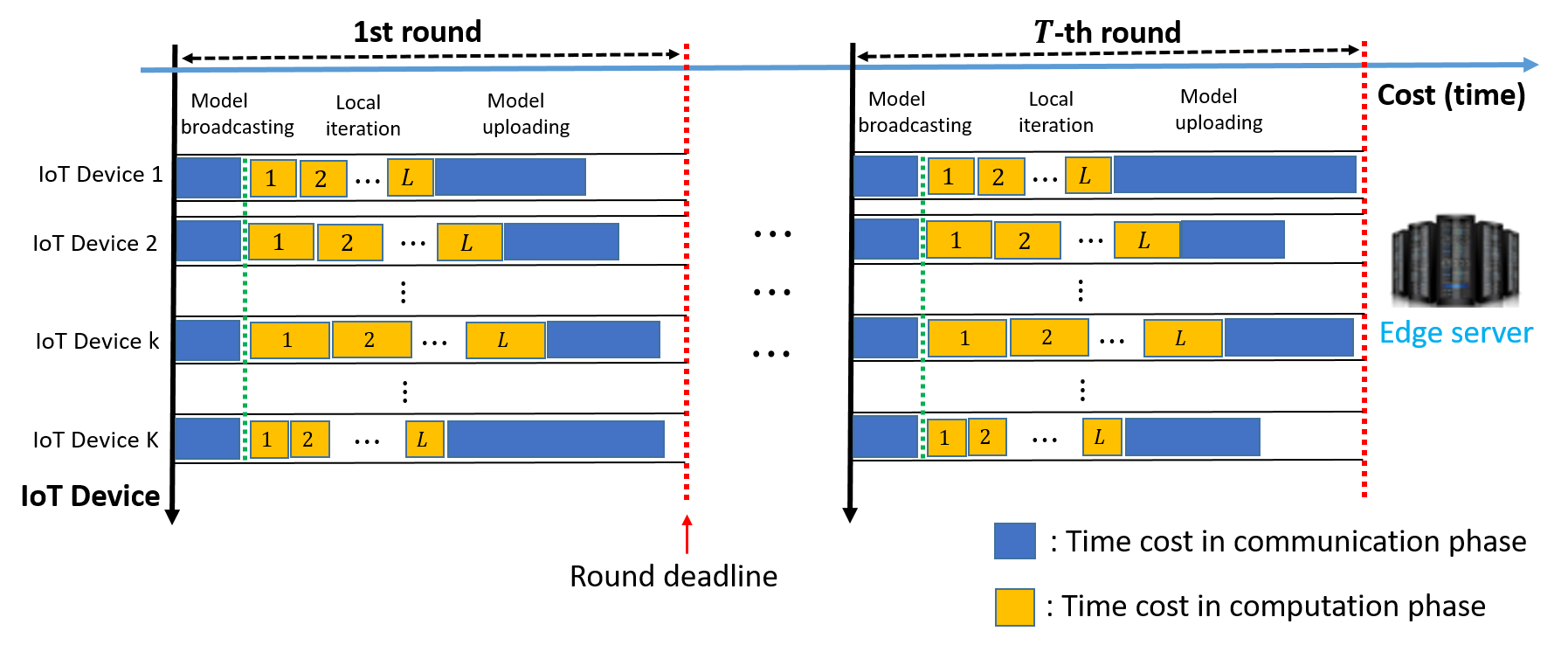}}
 \caption{The whole training process of FL, in each round, the time consumption of IoT device $k$ includes global model download, local model update and upload.}
 \label{fig:3-1}
 \end{figure*}

Fig.~\ref{fig:3-1} depicts the FL protocol, where the global model at the edge server and the local models at the IoT devices are trained in $T$ rounds. Each FL round is composed of resource request, IoT device selection, global model aggregation, global model download, local model update and upload. The edge server  initializes the hyperparameters of the global model, e.g., learning rate, batch size, and the weights of the global model.

\begin{itemize}

 \item \texttt{Resource Request}: The IoT devices send to the edge server the information needed, i.e., data size, for the device selection. 
 
  \item \texttt{IoT Device Selection}: The edge server selects the IoT devices to upload the local model for the training of the global model, where the details will be presented in the next section.  
 
 \item  \texttt{Global Model Aggregation}:  The edge server aggregates the local models of the selected IoT devices to produce  the global model.

\item \texttt{Global Model Download}: All the IoT devices download the the global model from the edge server.

\item  \texttt{Local Model Update and Upload}: 
The selected IoT devices individually train their local models according to  the FL parameters of the global model.
\end{itemize}

We consider $K$ number of IoT devices, where $k \in [1, K]$. Let   $x_{ki}$ and $y_{ki}$ denote the input (e.g., pixels of an image) and the output (e.g., labels of the image) of the FL model \cite{Adnan-2101-07511}, respectively. The dataset of device $k$ is denoted as  $\mathcal{D}_{t,k} = \{ \mathbf{x}_{ki}, y_{ki} \}_{i=1}^{D_{t,k}}$, where $D_{t,k}$ is the size of the dataset of device $k$ in the $t$-th round and data sample $i$ in device $k$. Let $f(\mathbf{w}, \mathbf{x}_{ki}, y_{ki})$ denote the loss function of FL, which captures  approximation errors over the input $\mathbf{x}_{ki}$ and the output $y_{ki}$. $\mathbf{w}$ is the weight parameter of the loss function of the neural network being trained according to the FL procedure. Given $D_{t,k}$, the loss function at IoT device $k$ can be specified as, 

\begin{equation}
   F_{t,k}(\mathbf{w}) = \frac{1}{D_{t,k}} \sum_{i = 1}^{D_{t,k}} f(\mathbf{w}, \mathbf{x}_{ki}, y_{ki}),
    \label{each-device-in-epoch-loss}
\end{equation}

where $ f(\mathbf{w}, \mathbf{x}_{ki}, y_{ki}) $ can be specified according to the FL structure. For example, $f(\mathbf{w}, \mathbf{x}_{ki}, y_{ki}) =  \frac{1}{2} (\mathbf{x}_{ki}^T \mathbf{w}- y_{ki})^2$  is used to model linear regression, or $f(\mathbf{w}, \mathbf{x}_{ki}, y_{ki}) = -\log(1 + \exp{(y_{ki} \mathbf{x}_{ki}^T \mathbf{w})})$  is for the model of  logistic regression \cite{shalev2014understanding}. Since the training of FL aims to minimize the weighted global loss function, we have

\begin{equation}
\begin{split}
   \mathop{\min}\limits_{\mathbf{w}}  F_t(\mathbf{w}) & =  \sum_{k=1}^{K} \frac{  D_{t,k}}{D_{t}} F_{t,k}(\mathbf{w})  \\ & = \frac{1}{D_t} \sum_{k=1}^{K} \sum_{i = 1}^{D_{t,k}} f(\mathbf{w}, \mathbf{x}_{ki}, y_{ki}), 
 \end{split}
\end{equation}
where  $D_{t} =  \sum_{k=1}^{K}  D_{t,k}$, it represents the total amount of data in the $t$-th round.

\subsection{Energy Model}
The energy consumption of the IoT devices accounts for the local training of the dataset and the transmissions of the local model. We assume that training a data sample at the IoT device  requires $c_k$ CPU cycles per bit. Given the data size of $D_{t,k}$, the number of CPU cycles for the local model training is  $c_{k}D_{t,k}$. We denote $f_k$ as the computation capacity of IoT device $k$, which is measured in CPU cycles per second. According to \cite{Dinh2021}, the computation time of training the local model at device $k$ in the each $t$-th round \cite{Dinh2021}, we have 

\begin{equation}
\tau_{t,k}^{train} = \frac{ c_{k} D_{t,k} L}{f_k}, \label{time-localtran}
\end{equation}
where $L$ is the number of local iterations of FL training.

According to \cite{yang2020energy} and \cite{mao2016dynamic}, the energy consumption on $c_{k}D_{t,k}$ CPU cycles at IoT device $k$ is
\begin{equation}
    E_{t,k1}^{cmp} = \zeta_k c_kD_{t,k} f_{k} ^2,
\end{equation}
where $\zeta_k $ is the effective capacitance coefficient of computing chipset for device $k$. To compute the local model, IoT device $k$ needs to compute $c_{k}D_{t,k}$ CPU cycles. Thus, the total computation energy at IoT device $k$ in the $t$-th round can be given as

\begin{equation}
    E_{t,k}^{cmp} =  L  E_{t,k1}^{cmp} = L \zeta_k c_kD_{t,k} f_{k}^2.
    \label{computation-energy}
\end{equation}

Furthermore, the achievable uplink transmit rate of IoT device $k$ is given by
\begin{equation}
r_{t,k}^{up} = b_{t,k} \log_{2} (1 + \frac{P_{t,k} G_{t,k}}{N_0 b_{t,k}}),
\label{transmitrate-up}
\end{equation}
where $b_{t,k}$ is the bandwidth allocated to IoT device $k$ by the edge server in the $t$-th round, $P_{t,k}$ is the power consumption of data transmit for IoT device $k$,  $G_{t,k}$ is the uplink channel gain between device $k$ and the edge server, and $N_0$ is the power spectral density of the Gaussian noise. 

The achievable downlink transmit rate of IoT device $k$ is
\begin{equation}
r_{t,k}^{down} = b_{t,k} \log_{2} (1 + \frac{P_{t}^{s} H_{t,k}}{N_0 b_{t,k}}),
\label{transmitrate-down}
\end{equation}
where $P_{t}^{s}$ denotes the transmit power of the edge server, and $H_{t,k}$ is the downlink wireless channel gain from the edge server to device $k$.

The downloading time of the global model at device $k$ in the $t$-th round is
\begin{equation}
  \tau_{t,k}^{down} = \frac{\mathfrak{S}_{d}}{ r_{t,k}^{down}},
 \label{time-download}
\end{equation}
where $ \mathfrak{S}_{d}$ denotes the  size of the global model. The transmission time of the local model at the $t$-th round can be given by 

\begin{equation}
  \tau_{t,k}^{up} = \frac{\mathfrak{S}_{u}}{ r_{t,k}^{up}}, \label{time-upload}
\end{equation}
where $\mathfrak{S}_{u}$ is the size of the local model. By substituting \eqref{transmitrate-up} to \eqref{time-upload}, the energy consumption of an IoT device on the local model transmission is

\begin{equation}
E_{t,k}^{up} = P_{t,k} \tau_{t,k}^{up}  = \frac{ P_{t,k} \mathfrak{S}_{u} }{b_{t,k} \log_{2} (1 + \frac{P_{t,k} G_{t,k}}{N_0b_{t,k}})}.
\label{energy-transmit}
\end{equation}

The total energy consumption $E_{t,k}^{c}$ of the selected IoT device $k$ in the $t$-th round is

\begin{equation}
E_{t,k}^{c} =  E_{t,k}^{cmp} + E_{t,k}^{up}
\label{energy-consumption-of-device}
\end{equation}

The remaining battery energy of the selected IoT device $k$ in the $t$-th round is
\begin{equation}
E_{t,k} = E_{t-1,k}- E_{t-1,k}^{c} + \Delta E_{t,k},
\end{equation}
where $\Delta E_{t,k}$ is the amount of harvested energy.


 \section{Problem Formulation}  \label{sec_problem_formu}
 In this section, we study the IoT device selection and transmit power allocation to maximize  the ratio of the learning accuracy of FL to the energy consumption of IoT devices.
 
 Let $\beta_{t,k}$ be a binary indicator. If IoT device $k$ is selected by the edge server in the $t$-th round, $\beta_{t,k} = 1$; otherwise, $\beta_{t,k} = 0$.  We define the accuracy of FL as the 
the fraction of predictions FL model got right. According to \cite{zhan2020learning, lim2020towards, khan2020federated}, the accuracy of FL, denoted by $\Gamma(\beta_{t,k})$,  can be simplified as, 
\begin{equation}
    \Gamma(\beta_{t,k} ) =   \log (1 +  \sum_{k=1}^{K} \mu_{k}\beta_{t,k} D_{t,k}) \quad \forall t \in \mathcal{T},
    \label{accuracy}
\end{equation}
where $\mu_{k} > 0$ is a system parameter ~\cite{lim2020towards}.
To improve the evenness of data size of the selected IoT devices, similar to \cite{Lyu2019}, we define the expectation of the difference between the total amount of data for all devices and the amount of data for the selected devices at the $t$-th training round, and normalize the expectation, we have,
\begin{equation}
   \Delta_t = \frac{\mathbb{E} \Big[\nu \sum_{k=1}^K D_{t,k}  - \beta_{t,k} D_{t,k}  \Big]}{\sum_{k=1}^K D_{t,k} } \quad \forall t \in \mathcal{T},  
\end{equation}
where $ \nu  \in (0, 1]$ is a constant value, the arrival data $D_{t, k}$ of device $k$ follows uniform distribution or normal distribution. The objective function can be determined as  $ \frac{\Gamma(\beta_{t,k})}{ \sum_{k=1}^{K} \beta_{t,k} E_{t,k}^c} - \Delta_t$, which aims to balance the learning accuracy of FL and energy consumption of the
IoT device and improve the evenness of data size of the selected IoT devices. We formulate the optimization  as \textbf{P1}. 
\begin{equation*}
\textbf{P1:} \quad \mathop{\max} \limits_{\beta_{t,k}, P_{t,k}} \   \sum_{t=1}^{T}  \bigg[     \frac{\Gamma(\beta_{t,k})}{ \sum_{k=1}^{K} \beta_{t,k} E_{t,k}^c} - \Delta_t \bigg]  
 \end{equation*}
\begin{equation}
  \mathbf{s.t.}:    \beta_{t,k}  E_{t,k}^c \leq  E_{t,k}, \ (t \in[1, T], k \in [1, K]). \label{energy-constr}
\end{equation}

\begin{equation}
    \Gamma(\beta_{t,k}) \geq \epsilon_0, \ ( \epsilon_0  \in (0, 1]).
     \label{con-accuracy}
\end{equation}

\begin{equation}
    \sum_{k=1}^{K} \beta_{t,k}  b_{t,k} \leq  B, \ (t \in[1, T], k \in [1, K]). 
    \label{con-bandwidth}
\end{equation}

\begin{equation}
    1 \leq \sum_{k=1}^{K} \beta_{t,k} \leq K, \ (t \in[1, T], k \in [1, K]). 
  \label{con-selection}
\end{equation}

\begin{equation}
\beta_{t,k} \tau_{t,k} \leq t_{d}, \ (t \in[1, T], k \in [1, K]). 
 \label{con-time}
\end{equation}


\begin{equation}
P_{k}^{min} \leq P_{t,k} \leq P_{k}^{max}, \ (t \in[1, T], k \in [1, K]).
\end{equation}

\begin{equation}
\beta_{t,k} \in \{0, 1\}, \ (t \in [1, T],  k \in [1, K]). \label{variable-constr}
\end{equation}

Specifically,  
\begin{itemize}
     \item Constraint ($\beta_{t,k}  E_{t,k}^c \leq  E_{t,k}$) guarantees that the selected IoT device has sufficient energy to complete the local model training in $t$.
     \item  Constraint ($\Gamma(\beta_{t,k}) \geq \epsilon_0 $) specifies the minimum requirement of the FL accuracy in $t$, where $\epsilon_0 \in (0, 1]$ defines the lower bound threshold. 
     \item  Constraint ($\sum_{k=1}^{K} \beta_{t,k}  b_{t,k} \leq  B$)  
     guarantees that the total bandwidth of the selected IoT devices is smaller than the bandwidth capacity $B$. 
     \item  Constraint ($1 \leq \sum_{k=1}^{K} \beta_{t,k} \leq K $) describes that at least one IoT device is selected for FL. 
    \item Constraint ($ \beta_{t,k} \tau_{t,k} \leq t_{d} $) ensures that the download, computation and transmit delay of the selected device has to be less than the duration of the round $t_{d}$. As shown in Fig.~\ref{fig:3-1}, the each $t$-th round contains the downloading time of the global model, and local computation and transmission time of the local model.  According to \eqref{time-localtran}, \eqref{time-download}, and \eqref{time-upload}, we can obtain the time delay $\tau_{t, k}$, i.e., $\tau_{t, k} = \tau_{t,k}^{train} + \tau_{t,k}^{up} + \tau_{t,k}^{down}$.  Note that the IoT device selection time and training time of the global model at the edge server can be neglected since the edge server supports more powerful CPUs than the IoT device.
    
     \item  Constraint ($P_{k}^{min} \leq P_{t,k} \leq P_{k}^{max}$)  indicates the upper and lower bounds of IoT devices' transmit power.
\end{itemize}

Problem \textbf{P1} involves nonlinear problems with continuous and integer variables \cite{sahinidis2019mixed}. The typical 0-1 Multidimensional Knapsack Problem (MKP)  \cite{kellerer2004multidimensional} is a special case of problem \textbf{P1}. Assume that transmit power is a constant. In this case, the only variable is the selected IoT device. The items to be put in the knapsack are the IoT devices with energy consumption $E_{t,k}$, data size $D_{t,k}$, and bandwidth $b_{t,k}$. The capacity of the knapsack is equal to the total bandwidth, the optimization variable $\beta_{t,k}$ is a binary indicator of item (IoT device) $k$ selection. $\beta_{t,k}$ is set to 1 to indicate that item $k$ is selected. Otherwise, $\beta_{t,k}$ is set to 0. The total reliability  of the knapsack has lower bounds which are equal to the minimum requirement of accuracy constraint \eqref{con-accuracy}. Note that every item to be put into the knapsack must obey the time constraint \eqref{con-time}. Coupled with the energy consumption of IoT devices on the transmission is a non-linear function of the transmit power, which has a continuous variable. Therefore, the proposed optimization is NP-hard. It is also mentioning that  problem \textbf{P1} involves  a long time horizon with  random and unpredictable data and energy arrivals. This leads to an intractable large state space of the problem require online optimization of selection and allocation decisions.

\section{DRL-based IoT Devices Selection and Resource Allocation} \label{sec_drl-based}
\subsection{POMDP Formulation for FL-based IoT Device Selection and Resource Allocation} \label{sec_pomdp} 
 The considered resource allocation can be formulated as a POMDP that is a generalization of a Markov decision process (MDP) with only partially observable states \cite{kaelbling1998planning}. A POMDP can be represented by a 6-tuple $(S, A, P, R, O, \Omega)$, where $S$ is the state space, $A$ is the action space, $P$ is the transition probability, $R$ is the reward function, $O$ is the observation space and $\Omega$ is the observation model. The edge server cannot observe the underlying state. Instead, an observation $S_{\alpha^{'}}^{o} \in O$ is received after a state transition to the next state $S_{\alpha^{'}}$ with the probability $\Omega (S_{\alpha^{'}}^{o} | S_{\alpha^{'}})$.

 \textit{State and Action Space}: According to problem \textbf{P1}, the state space of the POMDP consists of  data size, uplink channel gains, downlink channel gains, bandwidth and remaining energy of the IoT devices. The network state $S_{\alpha}$ is defined as
\begin{equation}  
  S_{\alpha} = \{(D_{\alpha, k}, G_{\alpha, k}, H_{\alpha, k}, b_{\alpha, k}, E_{\alpha, k}), k = 1, \cdots, K \}. 
\end{equation}

The action of the POMDP is the selection of the IoT devices for FL, denoted by  $\beta_{\alpha,k}$,  and the transmit powers of the selected IoT devices,  $P_{\alpha,k}$, as given by 
\begin{equation}
  \mathcal{A} \in \{ (\beta_{\alpha,k}, P_{\alpha,k}), k = 1, \cdots, K \},
\end{equation}
where $\beta_{\alpha, k} \in \{0, 1 \}$ and $P_{\alpha,k} \in [P_k^{min}, P_k^{max}]$.

\textit{Observation Space}: At each state  $S_{\alpha}$, the edge server can observe partially the network state from the selected IoT devices, where the state observation $S^o_{\alpha}$ can be packed in its uploaded local model. Particularly, the state observation $S_\alpha^o  \in S_{\alpha}$ is given by
\begin{equation}
  S_\alpha^o = \{(D_{\alpha, k}, G_{\alpha, k}, H_{\alpha, k}, b_{\alpha, k}, E_{\alpha, k})_{o}, k = 1, \cdots, K \}. 
\end{equation}
Note that the state of unselected IoT devices cannot be observed by the edge server.

\textit{Reward}: Let $ R(S_{\alpha^{'}}^o | S_\alpha^o, A_{\alpha} )$ denote the  immediate reward received when the action $A_{\alpha}  \in  \mathcal{A} $ is taken at state  $S_{\alpha}$.  The reward is defined to consist of the  AE gain that is the ratio of the FL accuracy to the energy consumption of the selected IoT devices, and the penalty resulting from the data unevenness among the selected IoT devices, i.e., 
  \begin{equation}
    R(S_{\alpha^{'}}^o | S_\alpha^o, A_{\alpha} ) =  \frac{\Gamma( \beta_{\alpha,k})}{ \sum_{k=1}^{K} \beta_{\alpha,k} E_{\alpha,k}^c}  - \Delta_{\alpha},  
  \end{equation}
where $   R(S_{\alpha^{'}}^o | S_\alpha^o, A_{\alpha} ) $ indicates that the state observation transits to subsequent $S_{\alpha^{'}}^o$  from the current state observation $S_\alpha^o$. For illustration convenience,  we use $R_{\alpha}$ to denote the reward in the following sections.

To evaluate the action selected by a policy $\pi_{\theta}$ with  parameters $\theta$, where $\pi_{\theta}$ is a mapping from state observations to actions, and the set of all policies is defined as $\Pi$. We aim to maximize the expected total reward denoted as action-value function $Q_{\pi_{\theta}}(S_\alpha^o, A_{\alpha})$,
\begin{equation}
    Q_{\pi_{\theta}}(S_\alpha^o, A_{\alpha}) = \mathop{\max}\limits_{\pi \in \Pi}  \mathbb{E}_{S_\alpha^o}^{\pi_{\theta}} \bigg\{  \sum_{n = 0}^{\infty}  \gamma^{n}  R_{\alpha} \bigg\},
    \label{action-value function}
\end{equation}
where $\gamma \in [0, 1]$ is a  discount factor for future state observations. $\mathbb{E}_{S_\alpha^o}^{\pi_{\theta}} \{\cdot\}$ takes the expectation with respect to policy $\pi_{\theta}$ and state observation $S_\alpha^o$.  According to the Bellman equation \cite{bellman1966dynamic}, the optimal action-value function \eqref{action-value function} of  a state-action pair $(S_\alpha^o, A_{\alpha})$ and the value of the subsequent state-action pair $( S_{\alpha^{'}}^o, A_{\alpha^{'}})$ can be further rewritten as 
\begin{equation}
\begin{split}
    Q_{\pi_{\theta}}(S_\alpha^o, A_{\alpha}) =  \mathop{\max}\limits_{\pi_{\theta} \in \Pi}  \mathbb{E}_{
    S_\alpha^o}^{\pi_{\theta}} \bigg\{ R_{\alpha} +  \gamma Q_{\pi_{\theta}}( S_{\alpha^{'}}^o, A_{\alpha^{'}})  \bigg\}.
\end{split}
\label{bellman-equation}
\end{equation}

The optimal action, namely, $A_{\alpha}^{*}$,  which satisfies \eqref{bellman-equation}, can be given by
\begin{equation}
\begin{split}
    A_{\alpha}^{*} =  \arg \mathop{\max}\limits_{\pi_{\theta} \in \Pi}  \mathbb{E}_{S_\alpha^o}^{\pi_{\theta}} \bigg\{ R_{\alpha} + \gamma Q_{\pi_{\theta}}( S_{\alpha^{'}}^o, A_{\alpha^{'}})  \bigg\},
\end{split}
\label{optimal-action}
\end{equation}
where $A_{\alpha}^{*}$ provides the maximized AE gain.

Given a practical scenario where the edge server has no prior
knowledge on the transition probabilities, we propose a FL-DLT3 framework that joint IoT device selection and transmit power allocation algorithm that utilizes TD3, one of the deep reinforcement learning techniques, to maximize the AE gain. 

The state and action space increases dramatically with the number of devices, and the action space has both continuous and discrete actions. Moreover, the network state, in practice, is not always observable at the edge server because the data size and remaining energy information of IoT devices are always kept locally before the server takes the devices selection. In view of these challenges, it is difficult to find the exact solution of \textbf{P1}. To this end, we develop FL-DLT3 framework to find the near optimal solution of problem \textbf{P1}. 

\begin{figure*}[htbp]
\centering
\includegraphics[scale=0.5]{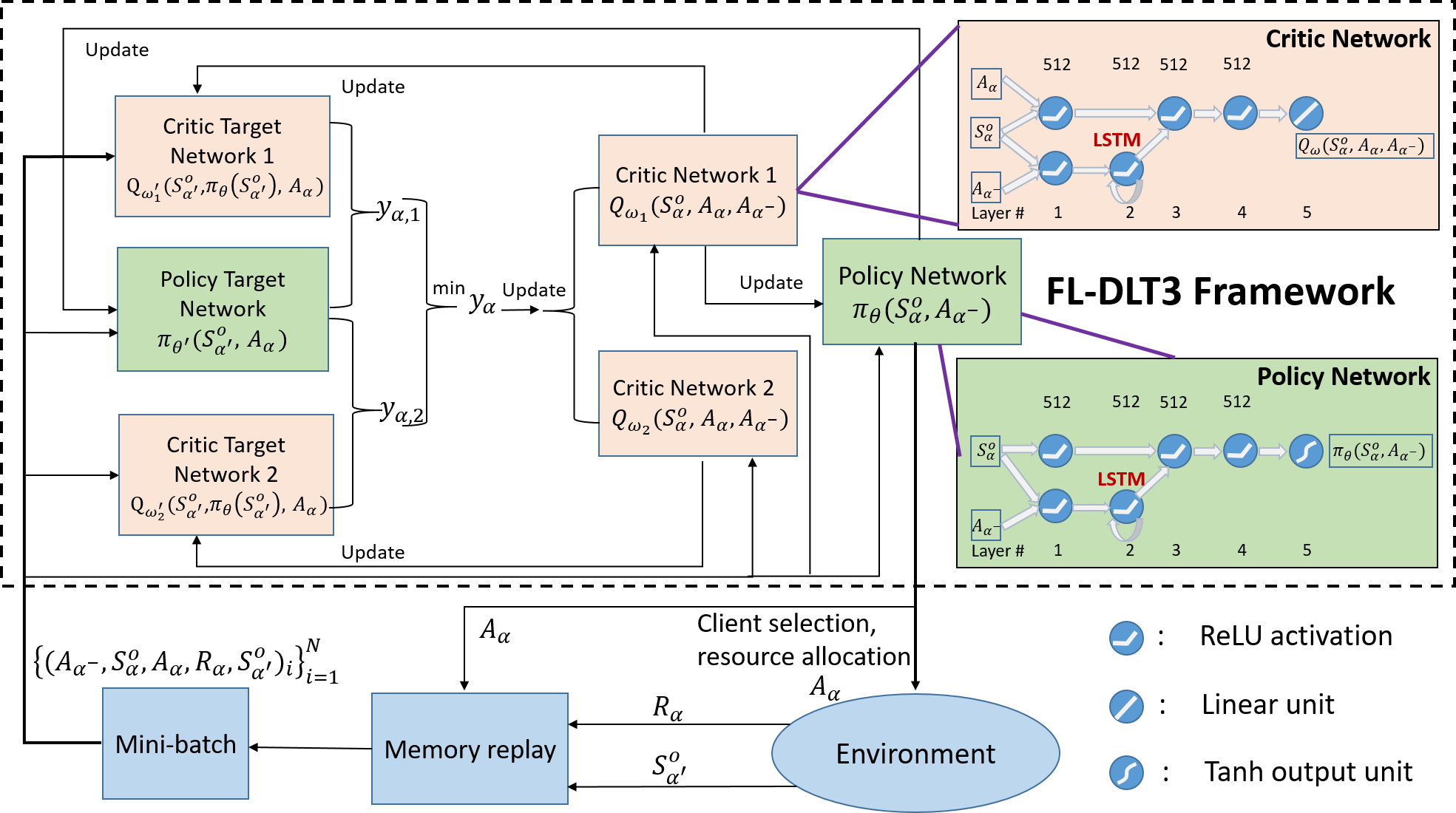}
 \caption{ An illustration of the proposed FL-DLT3 framework, where a policy network, a policy target network, two critic networks and two critic target networks are trained. Each of the networks consists of a feedforward branch and recurrent branch. A training round of FL is performed by the selected IoT devices with the allocated transmit power. Then, the edge server gets the reward $R_{\alpha}$ and the state update $S_{\alpha^{'}}^{o}$ of the IoT devices. The transition $\{(A_{\alpha^{-}},  S_\alpha^o, A_{\alpha},  R_\alpha, S_{\alpha^{'}}^o)\}$  is stored into the replay buffer $\mathcal{B}$, and a mini-batch of transitions is randomly sampled from $\mathcal{B}$ to train the policy and critic networks on the edge server.}
\label{FL-DLT3}
\end{figure*}

In general, FL-DLT3 consists of the TD3-based  deep reinforcement learning and the LSTM-based state characterization layer, as depicted in Fig.~\ref{FL-DLT3}. FL-DLT3 leverages the actor-critic neural network structure to develop the TD3-based  deep reinforcement learning \cite{fujimoto2018addressing}. The TD3 at the edge server is trained to optimize IoT device selection and transmit power allocation in a continuous action space, where the edge server has no priori information on the state transition probabilities. The LSTM is used to predict the hidden state of the IoT devices as input state of TD3. The AE gain is maximized over the large continuous state and action spaces. Specially, the global model is trained iteratively to improve the learning accuracy of the local model. With the growth of FL iterations,  FL-DLT3 optimally selects the IoT devices to maximize the accuracy of FL while ensuring the energy consumption requirement.

\subsection{TD3 on the Edge Server}
TD3 utilizes the deterministic policy gradient algorithm (DPG) \cite{silver2014deterministic} to optimally update the current policy by deterministically mapping network states to a specific action of the edge server. The critic is used to approximate the action-value function, which is related to the update of the actor, also known as policy. Moreover, the edge server stores the transition  about  previous actions of the edge server,
current observation, actions of the edge server, AE gain, next observations, i.e., $(A_{\alpha^{-}},  S_\alpha^o, A_{\alpha},  R_\alpha, S_{\alpha^{'}}^o)$,  into the replay buffer $\mathcal{B}$ at each training step. A mini-batch of $N$ transitions are randomly sampled from $\mathcal{B}$ to train the actor-critic networks. The action will be selected according to the policy network with exploration noise after the first $M$ time steps. Since the IoT device selection action output from the policy network may not  be binary, we classify them as binary by setting thresholds. Based on the TD3 framework, the policy and critic networks can be trained to approximate the optimal policy of the formulated POMDP problem.

For the continuous IoT device selection and transmit power allocation, the objective is to find the optimal policy $\pi_{\theta}$, which maximizes the expected AE gain $J(\theta) = \mathbb{E}_{S_\alpha^o}^{\pi_{\theta}}  \big[ R(S_0^o, A_0) \big] $, the parameterized policies $\pi_{\theta}$ can be updated by taking the gradient of the expected return with respect to $\theta$, i.e.,
\begin{equation}
    \nabla_{\theta} J(\theta) =  \mathbb{E}_{\pi_{\theta}}  \big[ \nabla_{A_{\alpha}} Q_{\pi_{\theta}}(S_\alpha^o, A_{\alpha})|_{A_{\alpha} = \pi(S_\alpha^o)} \nabla_{\theta} \pi_{\theta}(S_\alpha^o) \big],
    \label{update-theta}
\end{equation}
where the optimal action-value function can be approximated by the critic neural network \cite{mnih2015human} $Q_{\omega}(S_\alpha^o, A_{\alpha})$ with parameters $\omega$, which obtains the AE gain. To update $Q_{\omega}(S_\alpha^o, A_{\alpha})$, the critic neural
network minimizes the approximation loss between the
current target value and $Q_{\omega}(S_\alpha^o, A_{\alpha})$ by adjusting the parameter $\omega$,
\begin{equation}
\begin{split}
\mathop{\min}\limits_{\omega}  \mathbb{E} \Big[  \Big( R_{\alpha}  +   \gamma Q_{\omega^{'}} \big(S_{\alpha^{'}}^o, \pi_{\theta}(S_{\alpha^{'}}^o) \big) - Q_{\omega}(S_\alpha^o, A_{\alpha}) \Big)^2 \Big],
\end{split}
\end{equation}
where $\omega^{'}$ stands for the periodically updated parameters of the critic target network, $\pi_{\theta}(S_{\alpha^{'}}^o)$ is the policy target network takes action in the next state observation  $S_{\alpha^{'}}^o$.

However, the update of the critic neural network with the above value function may result in overestimation, which can produce sub-optimal policies of the policy network. To deal with the overestimation bias problem, TD3 uses two approximately independent critic networks $\{ Q_{\omega_1}, Q_{\omega_2} \}$ to estimate the value function, i.e., 
\begin{equation}
\begin{split}
   & y_{\alpha, 1} = R_{\alpha}  +  \gamma Q_{\omega_1^{'}}(S_{\alpha^{'}}^o, \pi_{\theta}(S_{\alpha^{'}}^o)),  \\
   & y_{\alpha, 2} = R_{\alpha}  +  \gamma Q_{\omega_2^{'}}(S_{\alpha^{'}}^o, \pi_{\theta}(S_{\alpha^{'}}^o)),
\end{split}
\end{equation}
where the minimum of the two estimators is utilized in the update of the value function, i.e., $y_{\alpha} = \min \{ y_{\alpha, 1}, y_{\alpha, 2} \}$.

Moreover, FL-DLT3 follows TD3 to update the policy and targets less frequently than the Q functions. Delayed policy updates consists of only updating the policy and  critic target network every $d$ intervals. And we also add noise to the target action, which makes the policy network less likely to exploit actions with high Q-value estimates. The proposed FL-DLT3 IoT device selection and transmit power allocation policy is shown in Algorithm \ref{FL-DLT3-algorithm}.   Since the network system has $K$ IoT devices, each state of the IoT devices is made up of the source of data size, uplink channel gain, downlink channel gain, bandwidth, and the remaining energy of the IoT device. It takes $T$ rounds of iterations to before the algorithm terminates.  Therefore, the complexity of FL-DLT3 is $\mathcal{O}\big( T[ 7K(n_{pa}+n_{pc1} + n_{pc2}) + (l_{pa}-1)n_{pa}^{2}  +  (l_{pc1}-1)n_{pc1}^{2} + (l_{pc2}-1)n_{pc2}^{2}  + 12 \times (7K\times N_{lstm} + (7K)^2 + 7K)] \big)$, where $n_{pa},n_{pc1}$ and $n_{pc2}$ are the number of neurons in the hidden layer of policy network, critic network 1 and critic network 2, respectively; $l_{pa},l_{pc1}$ and $l_{pc2}$ are the number of the hidden layers of policy network, critic network 1 and critic network 2, respectively; $N_{lstm}$ is the hidden size of LSTM.

\subsection{Network Architecture}
 Each network follows two-branch structure as in \cite{peng2018sim}, which consists of a feedforward branch and recurrent branch (as shown in  Fig.~\ref{FL-DLT3}). The feedforward branch and recurrent branch both  use five fully-connected layers neural network of 512 hidden nodes. The recurrent branch consists of an embedding layer of 512 fully connected units followed by 512 LSTM units. For each hidden layer (apart from the LSTM),  ReLU activations are used between the two hidden layers for both the policy and critic. The final tanh unit and linear unit following the output of the policy and critic, respectively.

\subsection{LSTM Layer}
The unknown network observation transitions resulting from time-varying data size, channel gains, bandwidth, and energy harvesting, which increase learning uncertainties and reduce learning accuracy. The edge server running the FL-DLT3 cannot observe the complete states of all the IoT devices.  It can only make the observation of a IoT device when the device is selected and uploads its state information to the edge server. The learning efficiency and accuracy of the TD3-based FL-DLT3 can be compromised by the imperfect knowledge of the states of the IoT devices. Motivated by this fact, we develop a state characterization layer to predict the states of the IoT devices which are not observable, and feed the predicted states into the every neural network of FL-DLT3.

For every policy (target) network, two critic (target) networks and memory replay, we use LSTM to predict their respective hidden states. The LSTM-based neural network is used to find out the hidden state in the environment, e.g.,  the state of a device that has not been selected previously. Moreover, the LSTM based state characterization layer helps to  accelerate the convergence of the FL. At time state $i$, the hidden states $h_{i}^{hid}$ is calculated by the following composite function
\begin{equation} 
h_{i}^{hid} = o_i  \tanh(C_i), 
\end{equation}
\begin{equation} 
o_{i} = \sigma( W_0 \cdot [C_i, h_{i-1}^{hid},  A_i ] + e_0), 
\end{equation}
\begin{equation} 
C_{i} = F_i C_{i-1} + p_i \tanh ( W_c \cdot [h_{i-1}^{hid},  A_i ] + e_c), 
\end{equation}
\begin{equation} 
F_{i} = \sigma ( W_f \cdot [h_{i-1}^{hid}, C_{i-1},  A_i ] + e_f), 
\end{equation}
\begin{equation} 
p_i = \sigma ( W_p \cdot [h_{i-1}^{hid}, C_{i-1},  A_i ] + e_p),
\end{equation}
where $o_{i}, C_{i}, F_{i}$, and $p_i$ denote the output gate, cell activation vectors, forget gate, and input gate of the LSTM layer, respectively.  $\sigma$, and $ \tanh$ refer to logistic sigmoid function and the hyperbolic tangent function, respectively. $\{ W_0, W_c,W_f,W_p \}$ are the weight matrix, and $\{ e_0, e_c,e_f,e_p \}$ are the bias matrix.

\begin{algorithm} 
\SetAlgoLined
	\caption{The developed FL-DLT3 IoT device selection and resource allocation policy}
  Initialize critic networks $\{ Q_{\omega_1}(S_\alpha^o, A_{\alpha}, A_{\alpha^{-}}),  Q_{\omega_2}(S_\alpha^o, A_{\alpha}, A_{\alpha^{-}}) \}$ and policy network $\pi_{\theta}(S_\alpha^o, A_{\alpha^{-}})$ with random parameters $\omega_1, \omega_2, \theta$. \\
  Initialize target networks $\omega_1 ^ {'} \leftarrow \omega_1, \omega_2 ^ {'} \leftarrow \omega_2, \theta^{'} \leftarrow \theta$. \\
  Initialize  environment $S_{\alpha}^o$, previous actions $A_{\alpha^{-}} \leftarrow 0 $, replay buffer $\mathcal{B}$.\\
  \For{$\alpha = 1,  \cdots, T$} 
	{
	\eIf{$\alpha \leq N$}{
	Explore $N$ steps with random policy, obtain $R_{\alpha}$ and $S_{\alpha^{'}}^o$, and storage the transition $(A_{\alpha^{-}}, S_\alpha^o, A_{\alpha}, R_{\alpha}, S_{\alpha^{'}}^o)$ of the $N$ steps into $\mathcal{B}$.
	}{
	Select action with exploration noise $A_{\alpha} \sim \pi_{\theta}(S_\alpha^o, A_{\alpha^{-}}) + \rho, \rho \sim \mathcal{N}(0, \sigma)$. \\
    Server allocates the selected IoT devices $P_{\alpha,k}$. \\
    Server performs model aggregation to obtain updated global model and distributes it to all selected devices in the next round. \\ %
    Server observes new observation $S_{\alpha^{'}}^o$, calculates $R_{\alpha}$, and stores the experience $(A_{\alpha^{-}}, S_\alpha^o, A_{\alpha}, R_{\alpha}, S_{\alpha^{'}}^o)$ into $\mathcal{B}$. \\
     Sample mini-batch of $N$ experiences $\{(A_{\alpha^{-}}, S_\alpha^o, A_{\alpha}, R_{\alpha}, S_{\alpha^{'}}^o)_i \}_{i=1}^N$ from $\mathcal{B}$.\\
    Obtain target action $\widehat{A}_{\alpha^{'}} \leftarrow \pi_{\theta^{'}}(S_{\alpha^{'}}^o, A_{\alpha} ) +  \widehat{\rho}, \widehat{\rho} \sim \rm{Clip}(\mathcal{N}(0, \widehat{\sigma}), -c, c )$. \\
$\widehat{y} \leftarrow R_{\alpha} + \gamma \mathop{\min} \limits_{m=1,2} Q_{\omega_m}(S_{\alpha^{'}}^o,\widehat{A}_{\alpha^{'}}, A_{\alpha})$. \\ 
Update critics $\omega_m \leftarrow \arg \mathop{\min} \limits_{\omega_m} \mathbb{E}(\widehat{y}  - Q_{\omega_m}(S_\alpha^o, A_{\alpha^{-}} ))^2$. \\
  \If{$\alpha$ mod $d$ = 0}  
  {
  Update $\theta$ by the deterministic policy gradient using equation \eqref{update-theta}. \\
Update target networks: \\
$ \omega_{m^{'}} \leftarrow  \varphi \omega_m + (1 - \varphi)\omega_{m^{'}}$. \\
 $ \theta^{'} \leftarrow \varphi \theta + (1 - \varphi) \theta^{'}$. \\
  }
	}
   }
	\label{FL-DLT3-algorithm}
\end{algorithm}

\section{Numerical Experiments} \label{section-numerical}
In this section, we present the implementation of the proposed 
FL-DLT3 on PyTorch, which is an open source machine learning library based on the Torch library. We compare FL-DLT3 with benchmarks in terms of network size, data size, and communication bandwidth. Table \ref{tab:experiment} specifies the configuration of simulation parameters.
\begin{table}[h!]
\small 
\begin{center}
    \caption{Simulation parameters}
    \label{tab:experiment}
    \begin{tabular}{l|l} 
     \hline
     \textbf{Parameters} & \textbf{Values} \\
      \hline
     Number of rounds ($T$) & 1000 \\
     Number of local iterations ($L$) & 4   \\
    \makecell[l]{For training one data sample \\ CPU  cycles per bit ($c_k$) } & 20 cycles / bit \\
    \makecell[l]{Computation capacity of IoT \\ device $k$ ($f_k$)}  & [2, 4] GHZ\\
     Transmit power of IoT device $k$ ($ P_{t,k}) $ &  [0.1, 60] W\\
     Transmit power of the edge server ($ P_{t}^{s}$) &  [100,1000] W\\
     Uplink channel gain  ($G_{t,k}$ ) & $[10^{-3}, 10^{-1}]$ dB\\
     Downlink channel gain ($H_{t,k}$ ) & $[10^{-1}, 10]$ dB\\ 
    \makecell[l]{Power spectral density of \\ the Gaussian noise ($ N_0$)}  & $1.0 \times 10^{-8}$ \\
     Parameter size of the global model ($S_d$) &  $ 1 \times 10^{4} $ bit \\
     Upload data size of IoT device  ($\mathfrak{S}_{u}$)  & $ 5 \times 10^{4} $ bit \\
     \makecell[l]{Binary indicator of device \\ selection ($\beta_{t,k}$ )} &  \{0, 1\} \\
     System parameter  ($\mu_{k}$) &  $4.2 \times 10^{-9}$\\
     Effective capacitance coefficient  ($\epsilon_k$) & $ 1.2 \times 10^{-28}$\\
     The amount of harvested energy ($\Delta E_{t,k}$) & [50, 200] J\\
     The constant value ($\nu$) & 1.0 \\
     
     Critic network learning rate  & $3 \times 10^{-4}$\\
     Policy network learning rate & $3 \times 10^{-4}$\\
     Neural network weight coefficient ($\phi$) & $5 \times 10^{-3}$ \\
    
    \makecell[l]{Interval of policy target \\ network update ($d$)}  & 10 \\
     Discount factor ($\gamma$) & 0.99 \\
     Batch size ($N$) & 45 \\ 
     Replay buffer size ($|\mathcal{B}|$) & $5 \times 10^{5}$ \\
     Exploration noise ($\sigma$) & $0.5$ \\
    Clipped normal noise ($c$) & $0.5$ \\
    \hline
    \end{tabular}
 \end{center}
\end{table}

\subsection{Implementation and Training of FL-DLT3}
We implement the proposed FL-DLT3 with Python 3.9. PyTorch is set up on a Linux workstation with 64-bit Ubuntu 18.04. FL-DLT3 trains the resource allocation with FL on 2 Nvidia's GPUs, one is GeForce GTX 1060 with 3 GB memory, the other is GeForce RTX 2060 with 6 GB memory. 

The experience replay memory can store  $5 \times 10^{5}$ training samples in terms of \textit{previous actions of the edge server, current observation, actions of the edge server, AE gain, next observations}. The previous actions and current observation are combined to feed into policy network and critic networks to infer the hidden state. Both state space and action space are updated by using Adam optimizer, where the learning rate is $3 \times 10^{-4}$. At one training step, the policy and critic networks are trained with a mini-batch of 45 transitions, sampled from the experience replay memory.

The policy target network is implemented by adding $\rho \sim \mathcal{N}(0, 0.5) $ to the actions chosen by the policy target network, clipped to (0, 60). The delayed policy updates the policy and  critic target networks every $d$ intervals, where $d = 10$. 

\subsection{AE Gain Performance}
For performance validation, we compare FL-DLT3 with existing state-of-the-art FL-based device scheduling approaches,  i.e., FedAECS \cite{zheng2021federated}, FedCS \cite{Nishio2019} and FedAvg \cite{McMahanMRA16}. 
\begin{itemize}
    \item \textbf{FedAECS:} The edge server selects the IoT devices to fulfill a predetermined ratio of FL accuracy to energy consumption, while meeting the requirement of accuracy and bandwidth. 
    \item \textbf{FedCS:}  Given the  bandwidth limit, the edge server selects  the maximum number of IoT devices for FL.
    \item  \textbf{FedAvg:}  Given the limit of the  bandwidth, the edge server determines the number of IoT devices for FL training, and randomly selects the IoT devices.
\end{itemize}

\begin{figure}[htbp]
\centering
\includegraphics[scale=0.5]{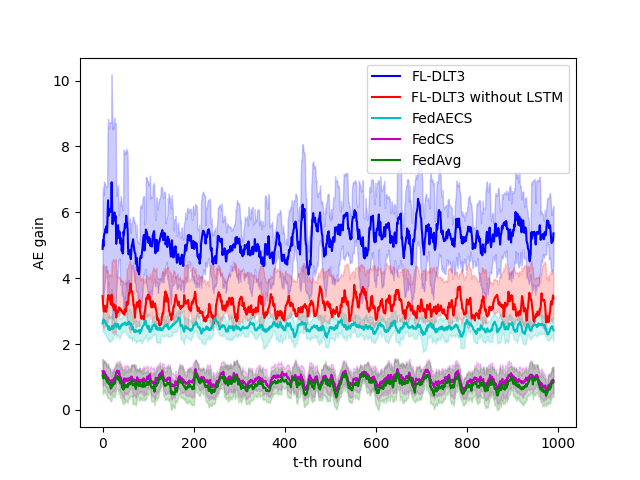}
\caption{Comparison of the AE gains, where $T = 1000$ and $K = 40$.}
\label{fig:experment-1}
\end{figure}

Fig. \ref{fig:experment-1} shows the AE gains, where the $t$-th round is from 1 to 1000 and $K$ = 40. The data size $D_{t,k}$ and bandwidth $b_{t,k}$ of the $K$ devices vary in [2, 10] MB and [10, 50] KHz, respectively. In general, the proposed FL-DLT3 achieves the highest AE gain, as compared to the existing FedAECS, FedCS, and FedAvg, improved by 51.8\%, 82.4\% and 85.0\% respectively.  The reason is that FL-DLT3 leverages experience replay and predict the states of the IoT devices which are not observable, however, FedAECS, FedCS, and
FedAvg are unable to predict the states of the IoT devices.

FL-DLT3 outperforms FL-DLT3 without the LSTM layer with a gain of 39.8\%. This is because the LSTM layer efficiently predicts the unknown network observation transitions, which enriches the training environment for FL and TD3. 

\begin{figure}[htbp]
\centering
\includegraphics[scale=0.55]{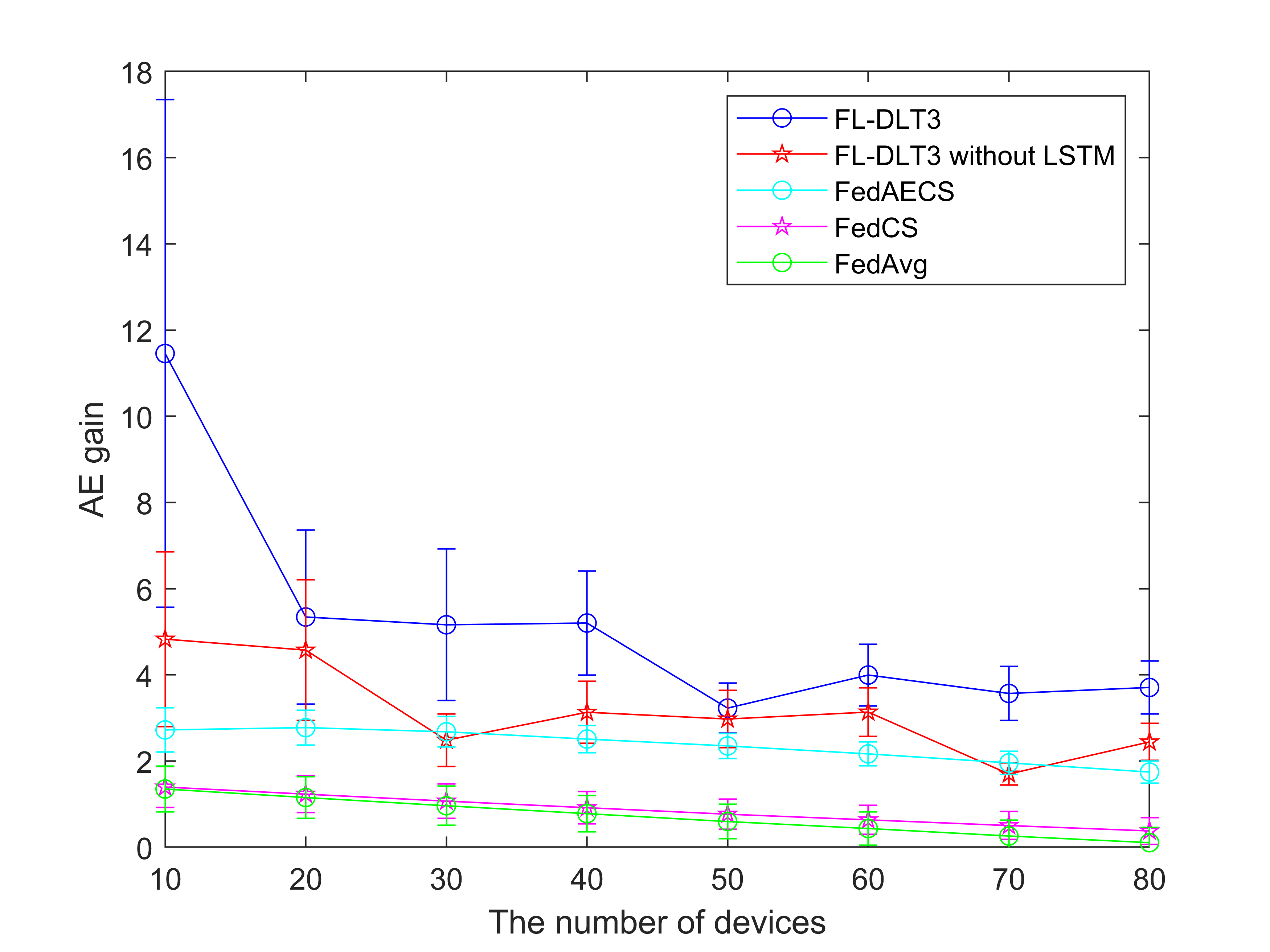}
\caption{Comparison of the AE gain obtained by using FL-DLT3, FL-DLT3 without LSTM, FedAECS, FedCs and FedAvg with different number of IoT devices}
\label{fig:experment-2}
\end{figure}

Fig.~\ref{fig:experment-2} studies the AE gain of the proposed FL-DLT3, where $K$ increases from 10 to 80. In general, the proposed FL-DLT3 achieves the highest AE gain   11.4557, as compared to the existing FedAECS (2.7212), FedCS (1.3952), and FedAvg (1.3498) given $K = 10$.  

\begin{figure}[htbp]
\centering
\includegraphics[scale=0.55]{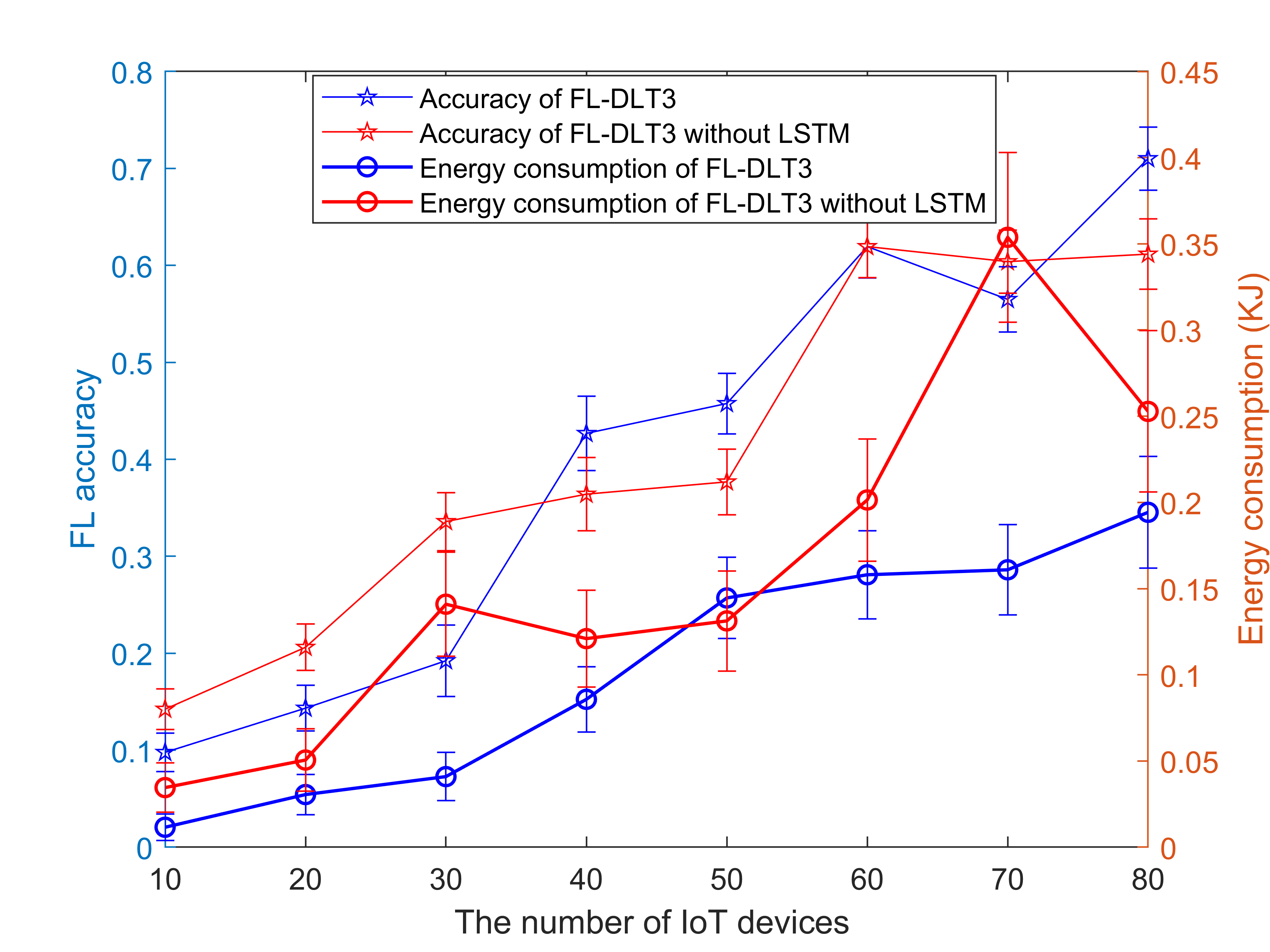}
\caption{Compare the FL accuracy and energy consumption of FL-DLT3 and FL-DLT3 without LSTM as the number of devices changes}
\label{fig:experment-2sub}
\end{figure}

In Fig.~\ref{fig:experment-2sub}, it can be observed that the energy consumption of FL-DLT3 dominates the performance when $K$ increase from 10 to 50. When $K >$   60, the FL accuracy dominates the performance. This confirms that the AE gain's fluctuation in Fig.~\ref{fig:experment-2} when K increase from 10 to 50.
The reasons is that  FL-DLT3 can select more IoT devices, hence the FL accuracy and energy consumption increase monotonically with the number of devices.

\begin{figure}[htbp]
\centering
\includegraphics[scale=0.5]{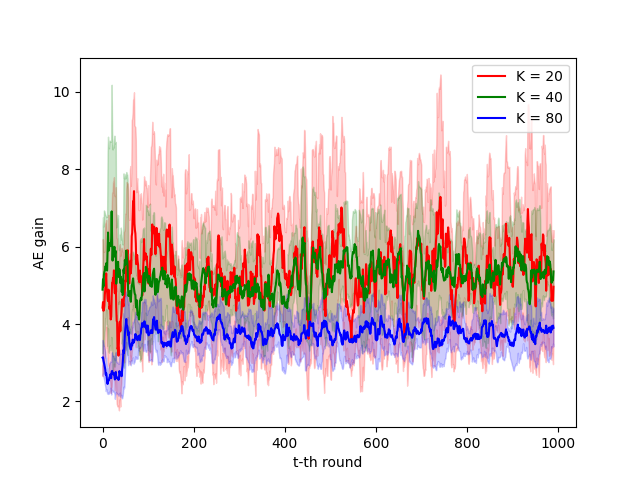}
\caption{The AE gain achieved by the proposed FL-DLT3 given $K$ =  20, 40, or 80.}
\label{fig:experment-3}
\end{figure}

Fig.~\ref{fig:experment-3} shows the AE gain  given 1000  rounds. With an increase of $K$, the AE gain achieved by FL-DLT3 decreases from 5.3411 to 3.7066. This also validates the performance in Fig. ~\ref{fig:experment-2}. 

\begin{figure}[htbp]
\centering
\includegraphics[scale=0.55]{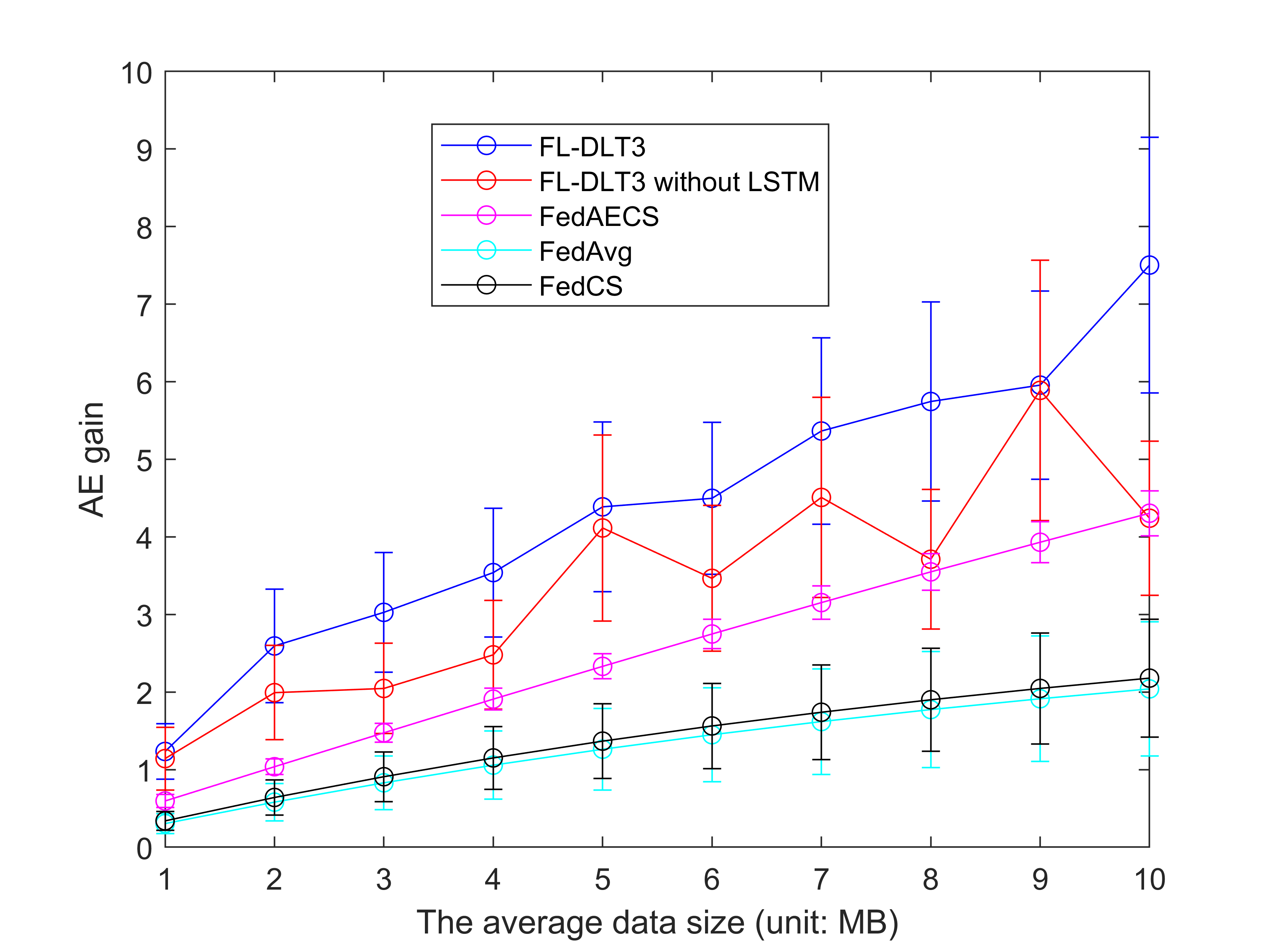}
\caption{AE gain varying with data size following normal distribution while keeping the same variance (0.2 MB) }
\label{fig:experment-4}
\end{figure}

 Fig.~\ref{fig:experment-4} depicts the AE gain of FL-DLT3 with regards to the average data size. In general, the AE gain increases with the growth of the data size. This is because the FL accuracy rises in a high rate while the energy consumption of the IoT devices on the training of FL-DLT3 slightly increases. Moreover, the AE gain obtained by FL-DLT3 is about twice that obtained by FedAECS, thanks to the transmit power allocation in FL-DLT3.

\begin{figure}[htbp]
\centering
\includegraphics[scale=0.55]{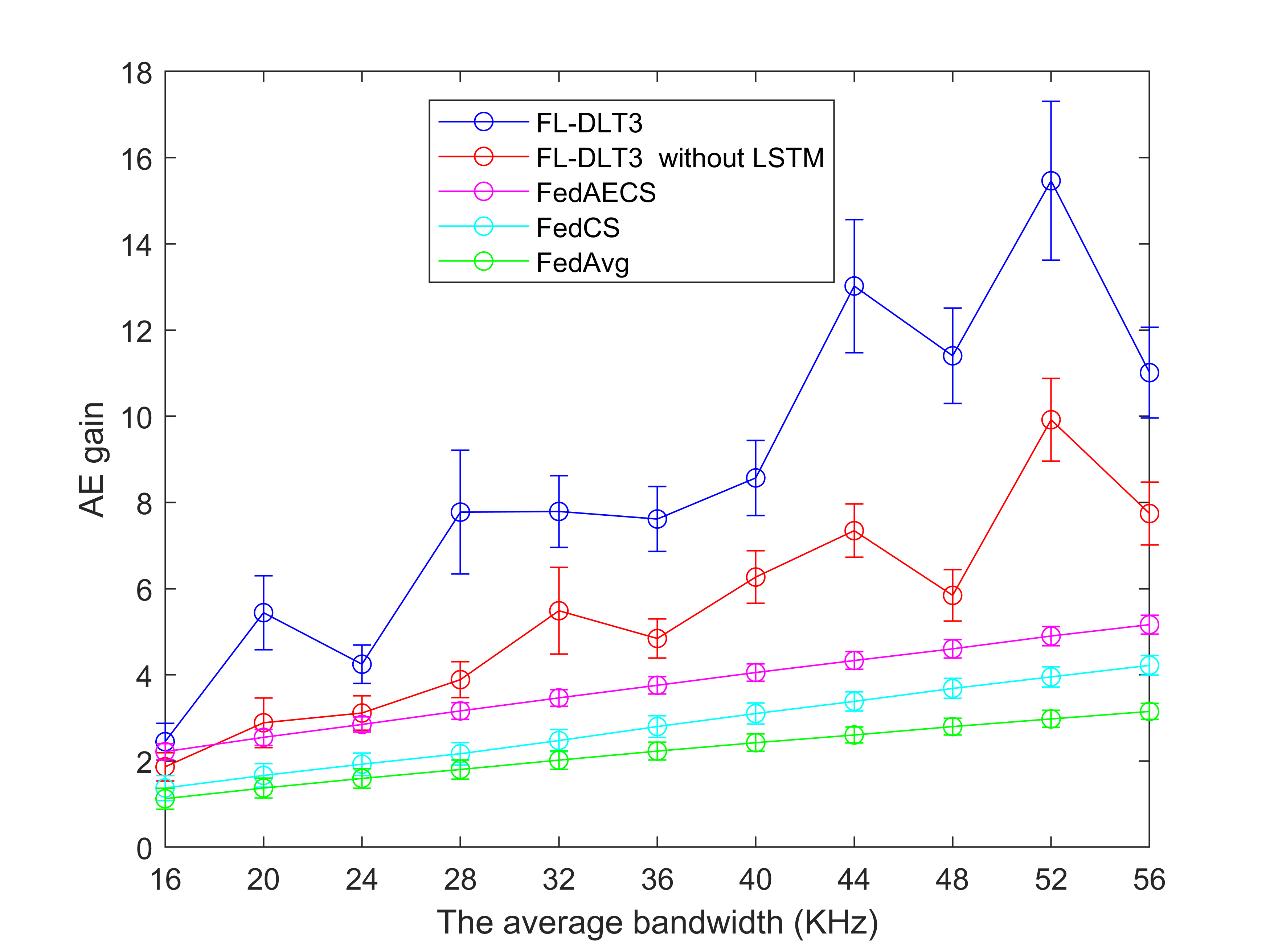}
\caption{AE gain varying with bandwidth following normal distribution while keeping the same variance (4 KHz) }
\label{fig:experment-5}
\end{figure}

 Fig.~\ref{fig:experment-5} describes the AE gain of FL-DLT3 with regards to the average bandwidth. Overall, the AE gain raises with an increase of the average bandwidth since the energy consumption of the IoT devices on the local models transmission is reduced,  in other words, a better channel quality leads to smaller packet retransmission and energy consumption.  In addition,  FL-DLT3 achieves the highest AE gain given different bandwidths. This is achieved by the LSTM layer is integrated into the FL-DLT3 to predict the time-varying bandwidth.

\begin{figure}[htbp]
\centering
\includegraphics[scale=0.55]{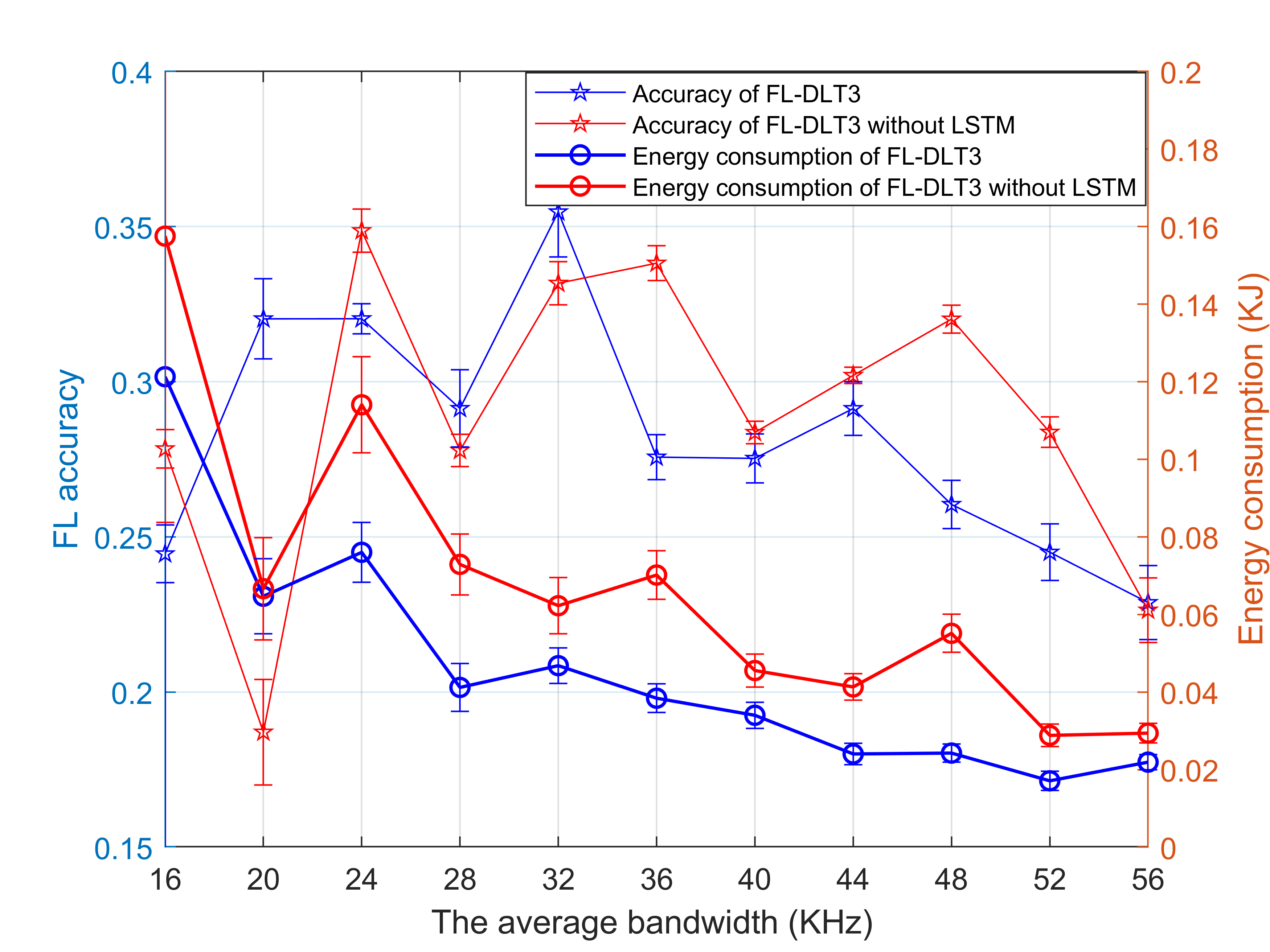}
\caption{Compare the FL accuracy and energy consumption of FL-DLT3 and FL-DLT3 without LSTM as the average channel size of IoT devices changes}
\label{fig:experment-5sub}
\end{figure}

Fig.~\ref{fig:experment-5sub} shows  the FL accuracy and energy consumption of FL-DLT3 and FL-DLT3 without LSTM in regard to the average bandwidth size of IoT devices changes. Generally, the energy consumption of FL-DLT3 decreases with an increase of the  bandwidth since the retransmission of the local models is reduced.

\begin{figure}[htbp]
\centering
\includegraphics[scale=0.55]{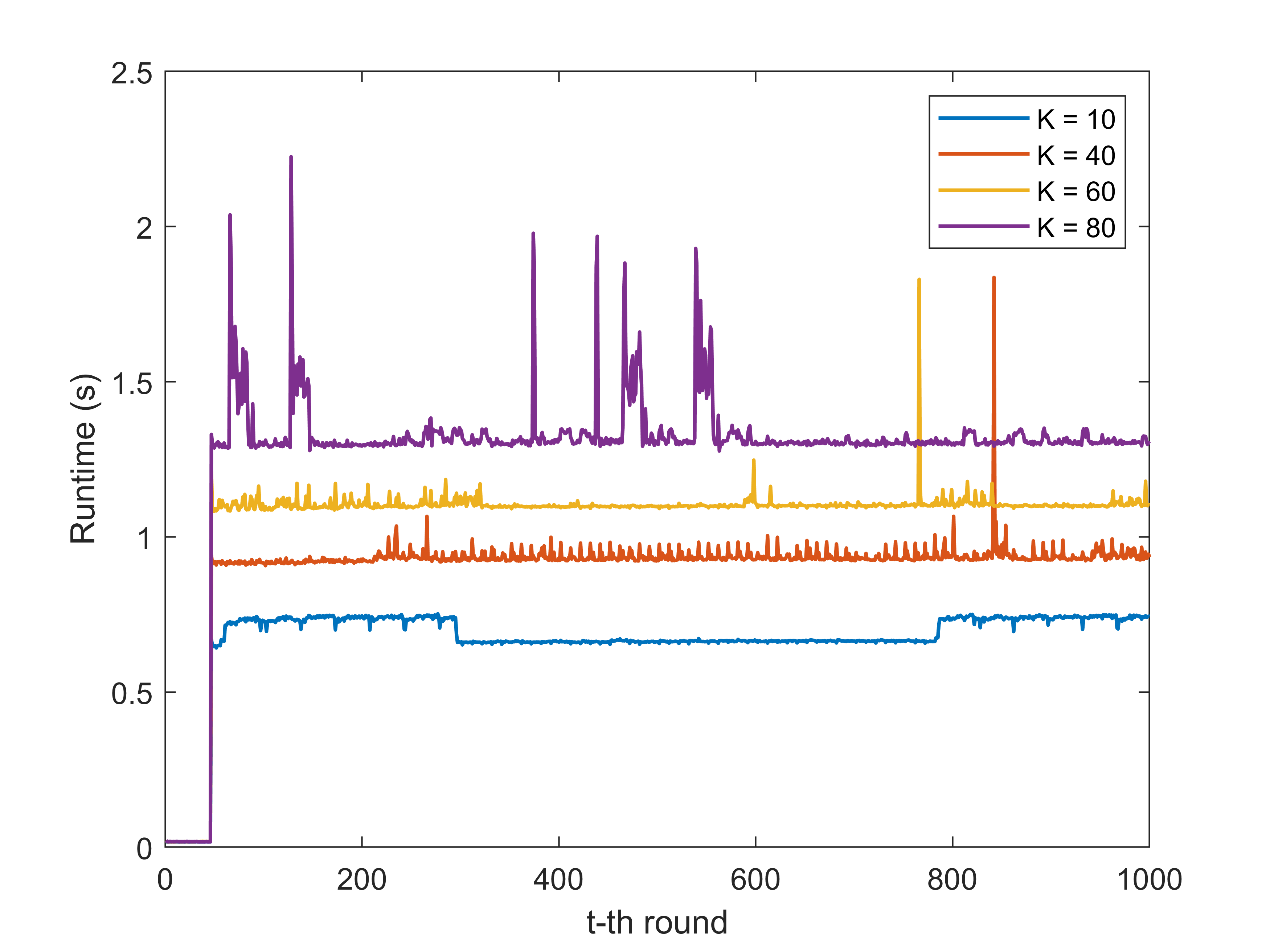}
\caption{Runtime with $t$-th round}
\label{fig:experment-6}
\end{figure}

 Fig.~\ref{fig:experment-6} shows the runtime of FL-DLT3, where $K$ is set to 10, 40, 60, or 80. In the first 45 rounds, the runtime of FL-DLT3 is about 0.018 s. This is because the experience replay buffer is initialized in which the training of FL-DLT3 is not conducted yet.  Once the learning experience is sufficient and FL-DLT3 carries out the training, the runtime raises to 0.74 s given $K$ = 10. When K increases to 80, the training of FL-DLT3 takes 1.3286 seconds. The reason is that an increase of the IoT devices results in a large state and action space, hence the learning time of FL-DLT3 increases.  In addition, the runtime of FL-DLT3 randomly fluctuates. This is due to random interruptions from other program executions that are concurrently operated on the server. 

\section{Conclusion} \label{sec_conclusion}
In this paper, we proposed FL-DLT3, which is a new deep reinforcement learning based resource allocation with FL for EdgeIoT. Given the large state and action space, FL-DLT3 learns the network state dynamics while maximizing the ratio of the learning accuracy of FL to the energy consumption of the IoT devices in a continuous domain. To  improve  the  learning  accuracy,  a  new  state  characterization  layer  based  on  LSTM  is  developed  in  FL-DLT3  to  predict  the time-varying  data  size,  bandwidth,  channel  gain,  and remaining energy of the IoT devices. FL-DLT3 is implemented in PyTorch. The effectiveness of FL-DLT3 is validated with the experimental data.

\section*{Acknowledgements}
This work was supported in part by the National Funds through FCT/MCTES (Portuguese Foundation for Science and Technology), within the CISTER Research Unit under Grant UIDP/UIDB/04234/2020, and in part by national funds through FCT, within project PTDC/EEI-COM/3362/2021 (ADANET).

\ifCLASSOPTIONcaptionsoff
  \newpage
\fi

\bibliographystyle{IEEEtran}
\bibliography{references} 

\begin{thebibliography}{10}
\providecommand{\url}[1]{#1}
\csname url@samestyle\endcsname
\providecommand{\newblock}{\relax}
\providecommand{\bibinfo}[2]{#2}
\providecommand{\BIBentrySTDinterwordspacing}{\spaceskip=0pt\relax}
\providecommand{\BIBentryALTinterwordstretchfactor}{4}
\providecommand{\BIBentryALTinterwordspacing}{\spaceskip=\fontdimen2\font plus
\BIBentryALTinterwordstretchfactor\fontdimen3\font minus
  \fontdimen4\font\relax}
\providecommand{\BIBforeignlanguage}[2]{{%
\expandafter\ifx\csname l@#1\endcsname\relax
\typeout{** WARNING: IEEEtran.bst: No hyphenation pattern has been}%
\typeout{** loaded for the language `#1'. Using the pattern for}%
\typeout{** the default language instead.}%
\else
\language=\csname l@#1\endcsname
\fi
#2}}
\providecommand{\BIBdecl}{\relax}
\BIBdecl

\bibitem{mao2017survey}
Y.~Mao, C.~You, J.~Zhang, K.~Huang, and K.~B. Letaief, ``A survey on mobile
  edge computing: The communication perspective,'' \emph{IEEE Communications
  Surveys \& Tutorials}, vol.~19, no.~4, pp. 2322--2358, 2017.

\bibitem{sun2016edgeiot}
X.~Sun and N.~Ansari, ``Edgeiot: Mobile edge computing for the internet of
  things,'' \emph{IEEE Communications Magazine}, vol.~54, no.~12, pp. 22--29,
  2016.

\bibitem{Deng9001216}
S.~Deng, Z.~Xiang, P.~Zhao, J.~Taheri, H.~Gao, J.~Yin, and A.~Y. Zomaya,
  ``Dynamical resource allocation in edge for trustable internet-of-things
  systems: A reinforcement learning method,'' \emph{IEEE Transactions on
  Industrial Informatics}, vol.~16, no.~9, pp. 6103--6113, 2020.

\bibitem{mach2017mobile}
P.~Mach and Z.~Becvar, ``Mobile edge computing: A survey on architecture and
  computation offloading,'' \emph{IEEE Communications Surveys \& Tutorials},
  vol.~19, no.~3, pp. 1628--1656, 2017.

\bibitem{9352033}
O.~A. Wahab, A.~Mourad, H.~Otrok, and T.~Taleb, ``Federated machine learning:
  Survey, multi-level classification, desirable criteria and future directions
  in communication and networking systems,'' \emph{IEEE Communications Surveys
  Tutorials}, vol.~23, no.~2, pp. 1342--1397, 2021.

\bibitem{Liangxiao}
\BIBentryALTinterwordspacing
L.~Xiao, X.~Wan, C.~Dai, X.~Du, X.~Chen, and M.~Guizani, ``Security in mobile
  edge caching with reinforcement learning,'' \emph{{IEEE} Wirel. Commun.},
  vol.~25, no.~3, pp. 116--122, 2018. [Online]. Available:
  \url{https://doi.org/10.1109/MWC.2018.1700291}
\BIBentrySTDinterwordspacing

\bibitem{9183799}
S.~A. Rahman, H.~Tout, C.~Talhi, and A.~Mourad, ``Internet of things intrusion
  detection: Centralized, on-device, or federated learning?'' \emph{IEEE
  Network}, vol.~34, no.~6, pp. 310--317, 2020.

\bibitem{mcmahan2017communication}
B.~McMahan, E.~Moore, D.~Ramage, S.~Hampson, and B.~A. y~Arcas,
  ``Communication-efficient learning of deep networks from decentralized
  data,'' in \emph{Artificial Intelligence and Statistics}.\hskip 1em plus
  0.5em minus 0.4em\relax PMLR, 2017, pp. 1273--1282.

\bibitem{xiao2018iot}
L.~Xiao, X.~Wan, X.~Lu, Y.~Zhang, and D.~Wu, ``Iot security techniques based on
  machine learning: How do iot devices use ai to enhance security?'' \emph{IEEE
  Signal Processing Magazine}, vol.~35, no.~5, pp. 41--49, 2018.

\bibitem{Ji2019}
S.~Ji, S.~Pan, G.~Long, X.~Li, J.~Jiang, and Z.~Huang, ``Learning private
  neural language modeling with attentive aggregation,'' in \emph{2019
  International Joint Conference on Neural Networks (IJCNN)}, 2019, pp. 1--8.

\bibitem{LiYijing2021}
Y.~Li, X.~Tao, X.~Zhang, J.~Liu, and J.~Xu, ``Privacy-preserved federated
  learning for autonomous driving,'' \emph{IEEE Transactions on Intelligent
  Transportation Systems}, pp. 1--12, 2021.

\bibitem{McMahanMRA16}
\BIBentryALTinterwordspacing
H.~B. McMahan, E.~Moore, D.~Ramage, and B.~A. y~Arcas, ``Federated learning of
  deep networks using model averaging,'' \emph{CoRR}, vol. abs/1602.05629,
  2016. [Online]. Available: \url{http://arxiv.org/abs/1602.05629}
\BIBentrySTDinterwordspacing

\bibitem{kang2019}
J.~Kang, Z.~Xiong, D.~Niyato, S.~Xie, and J.~Zhang, ``Incentive mechanism for
  reliable federated learning: A joint optimization approach to combining
  reputation and contract theory,'' \emph{IEEE Internet of Things Journal},
  vol.~6, no.~6, pp. 10\,700--10\,714, 2019.

\bibitem{Lim2021}
W.~Y.~B. Lim, J.~S. Ng, Z.~Xiong, D.~Niyato, C.~Miao, and D.~I. Kim, ``Dynamic
  edge association and resource allocation in self-organizing hierarchical
  federated learning networks,'' \emph{IEEE Journal on Selected Areas in
  Communications}, vol.~39, no.~12, pp. 3640--3653, 2021.

\bibitem{ji2019learning}
J.~Ji, X.~Chen, Q.~Wang, L.~Yu, and P.~Li, ``Learning to learn gradient
  aggregation by gradient descent.'' in \emph{IJCAI}, 2019, pp. 2614--2620.

\bibitem{9220780}
S.~Abdulrahman, H.~Tout, H.~Ould-Slimane, A.~Mourad, C.~Talhi, and M.~Guizani,
  ``A survey on federated learning: The journey from centralized to distributed
  on-site learning and beyond,'' \emph{IEEE Internet of Things Journal},
  vol.~8, no.~7, pp. 5476--5497, 2021.

\bibitem{Dinesh}
\BIBentryALTinterwordspacing
D.~C. Verma, S.~B. Calo, S.~Witherspoon, E.~Bertino, A.~A. Jabal, A.~Swami,
  G.~Cirincione, S.~Julier, G.~White, G.~de~Mel, and G.~Pearson, ``Federated
  learning for coalition operations,'' \emph{CoRR}, vol. abs/1910.06799, 2019.
  [Online]. Available: \url{http://arxiv.org/abs/1910.06799}
\BIBentrySTDinterwordspacing

\bibitem{Nishio2019}
T.~Nishio and R.~Yonetani, ``Client selection for federated learning with
  heterogeneous resources in mobile edge,'' in \emph{ICC 2019 - 2019 IEEE
  International Conference on Communications (ICC)}, 2019, pp. 1--7.

\bibitem{Shunfeng2021}
\BIBentryALTinterwordspacing
S.~Chu, J.~Li, J.~Wang, Z.~Wang, M.~Ding, Y.~Zang, Y.~Qian, and W.~Chen,
  ``Federated learning over wireless channels: Dynamic resource allocation and
  task scheduling,'' \emph{CoRR}, vol. abs/2106.06934, 2021. [Online].
  Available: \url{https://arxiv.org/abs/2106.06934}
\BIBentrySTDinterwordspacing

\bibitem{AnhLNKW19}
\BIBentryALTinterwordspacing
T.~T. Anh, N.~C. Luong, D.~Niyato, D.~I. Kim, and L.~Wang, ``Efficient training
  management for mobile crowd-machine learning: {A} deep reinforcement learning
  approach,'' \emph{{IEEE} Wirel. Commun. Lett.}, vol.~8, no.~5, pp.
  1345--1348, 2019. [Online]. Available:
  \url{https://doi.org/10.1109/LWC.2019.2917133}
\BIBentrySTDinterwordspacing

\bibitem{wang2019adaptive}
S.~Wang, T.~Tuor, T.~Salonidis, K.~K. Leung, C.~Makaya, T.~He, and K.~Chan,
  ``Adaptive federated learning in resource constrained edge computing
  systems,'' \emph{IEEE Journal on Selected Areas in Communications}, vol.~37,
  no.~6, pp. 1205--1221, 2019.

\bibitem{Wahab2021}
S.~Abdulrahman, H.~Tout, A.~Mourad, and C.~Talhi, ``Fedmccs: Multicriteria
  client selection model for optimal iot federated learning,'' \emph{IEEE
  Internet of Things Journal}, vol.~8, no.~6, pp. 4723--4735, 2021.

\bibitem{yoshida2020mab}
N.~Yoshida, T.~Nishio, M.~Morikura, and K.~Yamamoto, ``Mab-based client
  selection for federated learning with uncertain resources in mobile
  networks,'' in \emph{2020 IEEE Globecom Workshops (GC Wkshps}.\hskip 1em plus
  0.5em minus 0.4em\relax IEEE, 2020, pp. 1--6.

\bibitem{xu2021online}
B.~Xu, W.~Xia, J.~Zhang, T.~Q. Quek, and H.~Zhu, ``Online client scheduling for
  fast federated learning,'' \emph{IEEE Wireless Communications Letters}, 2021.

\bibitem{xia2020}
W.~Xia, T.~Q.~S. Quek, K.~Guo, W.~Wen, H.~H. Yang, and H.~Zhu, ``Multi-armed
  bandit-based client scheduling for federated learning,'' \emph{IEEE
  Transactions on Wireless Communications}, vol.~19, no.~11, pp. 7108--7123,
  2020.

\bibitem{zhu2021federated}
Z.~Zhu, S.~Wan, P.~Fan, and K.~B. Letaief, ``Federated multiagent actor--critic
  learning for age sensitive mobile-edge computing,'' \emph{IEEE Internet of
  Things Journal}, vol.~9, no.~2, pp. 1053--1067, 2021.

\bibitem{zhan9139873}
Y.~Zhan, P.~Li, and S.~Guo, ``Experience-driven computational resource
  allocation of federated learning by deep reinforcement learning,'' in
  \emph{2020 IEEE International Parallel and Distributed Processing Symposium
  (IPDPS)}, 2020, pp. 234--243.

\bibitem{kwon9067847}
D.~Kwon, J.~Jeon, S.~Park, J.~Kim, and S.~Cho, ``Multiagent ddpg-based deep
  learning for smart ocean federated learning iot networks,'' \emph{IEEE
  Internet of Things Journal}, vol.~7, no.~10, pp. 9895--9903, 2020.

\bibitem{MeixiaTao}
B.~Yin, Z.~Chen, and M.~Tao, ``Joint user scheduling and resource allocation
  for federated learning over wireless networks,'' in \emph{GLOBECOM 2020 -
  2020 IEEE Global Communications Conference}, 2020, pp. 1--6.

\bibitem{zheng2021federated}
J.~Zheng, K.~Li, E.~Tovar, and M.~Guizani, ``Federated learning for
  energy-balanced client selection in mobile edge computing,'' in \emph{2021
  International Wireless Communications and Mobile Computing (IWCMC)}.\hskip
  1em plus 0.5em minus 0.4em\relax IEEE, 2021, pp. 1942--1947.

\bibitem{Adnan-2101-07511}
\BIBentryALTinterwordspacing
A.~Qayyum, K.~Ahmad, M.~A. Ahsan, A.~I. Al{-}Fuqaha, and J.~Qadir,
  ``Collaborative federated learning for healthcare: Multi-modal {COVID-19}
  diagnosis at the edge,'' \emph{CoRR}, vol. abs/2101.07511, 2021. [Online].
  Available: \url{https://arxiv.org/abs/2101.07511}
\BIBentrySTDinterwordspacing

\bibitem{shalev2014understanding}
S.~Shalev-Shwartz and S.~Ben-David, \emph{Understanding machine learning: From
  theory to algorithms}.\hskip 1em plus 0.5em minus 0.4em\relax Cambridge
  university press, 2014.

\bibitem{Dinh2021}
\BIBentryALTinterwordspacing
C.~T. Dinh, N.~H. Tran, M.~N.~H. Nguyen, C.~S. Hong, W.~Bao, A.~Y. Zomaya, and
  V.~Gramoli, ``Federated learning over wireless networks: Convergence analysis
  and resource allocation,'' \emph{IEEE/ACM Trans. Netw.}, vol.~29, no.~1, p.
  398–409, feb 2021. [Online]. Available:
  \url{https://doi.org/10.1109/TNET.2020.3035770}
\BIBentrySTDinterwordspacing

\bibitem{yang2020energy}
Z.~Yang, M.~Chen, W.~Saad, C.~S. Hong, and M.~Shikh-Bahaei, ``Energy efficient
  federated learning over wireless communication networks,'' \emph{IEEE
  Transactions on Wireless Communications}, vol.~20, no.~3, pp. 1935--1949,
  2020.

\bibitem{mao2016dynamic}
Y.~Mao, J.~Zhang, and K.~B. Letaief, ``Dynamic computation offloading for
  mobile-edge computing with energy harvesting devices,'' \emph{IEEE Journal on
  Selected Areas in Communications}, vol.~34, no.~12, pp. 3590--3605, 2016.

\bibitem{zhan2020learning}
Y.~Zhan, P.~Li, Z.~Qu, D.~Zeng, and S.~Guo, ``A learning-based incentive
  mechanism for federated learning,'' \emph{IEEE Internet of Things Journal},
  2020.

\bibitem{lim2020towards}
W.~Y.~B. Lim, J.~Huang, Z.~Xiong, J.~Kang, D.~Niyato, X.-S. Hua, C.~Leung, and
  C.~Miao, ``Towards federated learning in uav-enabled internet of vehicles: A
  multi-dimensional contract-matching approach,'' \emph{IEEE Transactions on
  Intelligent Transportation Systems}, 2021.

\bibitem{khan2020federated}
L.~U. Khan, S.~R. Pandey, N.~H. Tran, W.~Saad, Z.~Han, M.~N. Nguyen, and C.~S.
  Hong, ``Federated learning for edge networks: Resource optimization and
  incentive mechanism,'' \emph{IEEE Communications Magazine}, vol.~58, no.~10,
  pp. 88--93, 2020.

\bibitem{Lyu2019}
X.~Lyu, C.~Ren, W.~Ni, H.~Tian, R.~P. Liu, and E.~Dutkiewicz, ``Optimal online
  data partitioning for geo-distributed machine learning in edge of wireless
  networks,'' \emph{IEEE Journal on Selected Areas in Communications}, vol.~37,
  no.~10, pp. 2393--2406, 2019.

\bibitem{sahinidis2019mixed}
N.~V. Sahinidis, ``Mixed-integer nonlinear programming 2018,'' 2019.

\bibitem{kellerer2004multidimensional}
H.~Kellerer, U.~Pferschy, and D.~Pisinger, ``Multidimensional knapsack
  problems,'' in \emph{Knapsack problems}.\hskip 1em plus 0.5em minus
  0.4em\relax Springer, 2004, pp. 235--283.

\bibitem{kaelbling1998planning}
L.~P. Kaelbling, M.~L. Littman, and A.~R. Cassandra, ``Planning and acting in
  partially observable stochastic domains,'' \emph{Artificial intelligence},
  vol. 101, no. 1-2, pp. 99--134, 1998.

\bibitem{bellman1966dynamic}
R.~Bellman, ``Dynamic programming,'' \emph{Science}, vol. 153, no. 3731, pp.
  34--37, 1966.

\bibitem{fujimoto2018addressing}
S.~Fujimoto, H.~Hoof, and D.~Meger, ``Addressing function approximation error
  in actor-critic methods,'' in \emph{International Conference on Machine
  Learning}.\hskip 1em plus 0.5em minus 0.4em\relax PMLR, 2018, pp. 1587--1596.

\bibitem{silver2014deterministic}
D.~Silver, G.~Lever, N.~Heess, T.~Degris, D.~Wierstra, and M.~Riedmiller,
  ``Deterministic policy gradient algorithms,'' in \emph{International
  conference on machine learning}.\hskip 1em plus 0.5em minus 0.4em\relax PMLR,
  2014, pp. 387--395.

\bibitem{mnih2015human}
V.~Mnih, K.~Kavukcuoglu, D.~Silver, A.~A. Rusu, J.~Veness, M.~G. Bellemare,
  A.~Graves, M.~Riedmiller, A.~K. Fidjeland, G.~Ostrovski \emph{et~al.},
  ``Human-level control through deep reinforcement learning,'' \emph{nature},
  vol. 518, no. 7540, pp. 529--533, 2015.

\bibitem{peng2018sim}
X.~B. Peng, M.~Andrychowicz, W.~Zaremba, and P.~Abbeel, ``Sim-to-real transfer
  of robotic control with dynamics randomization,'' in \emph{2018 IEEE
  international conference on robotics and automation (ICRA)}.\hskip 1em plus
  0.5em minus 0.4em\relax IEEE, 2018, pp. 3803--3810.

\end{thebibliography}

\begin{IEEEbiography}[{\includegraphics[width=1in,height=1.25in,clip,keepaspectratio]{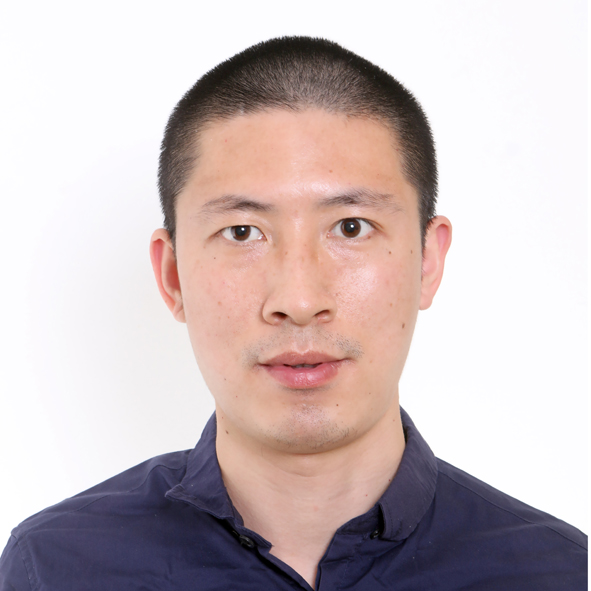}}]{Jingjing Zheng}
is a Student Researcher at CISTER Research Centre and is currently pursuing the Ph.D. degree in Electrical and Computer Engineering at the University of Porto, Porto, Portugal.
His main research interests include federated learning,
edge computing, and IoT security.\end{IEEEbiography}

\begin{IEEEbiography}[{\includegraphics[width=1in,height=1.25in,clip,keepaspectratio]{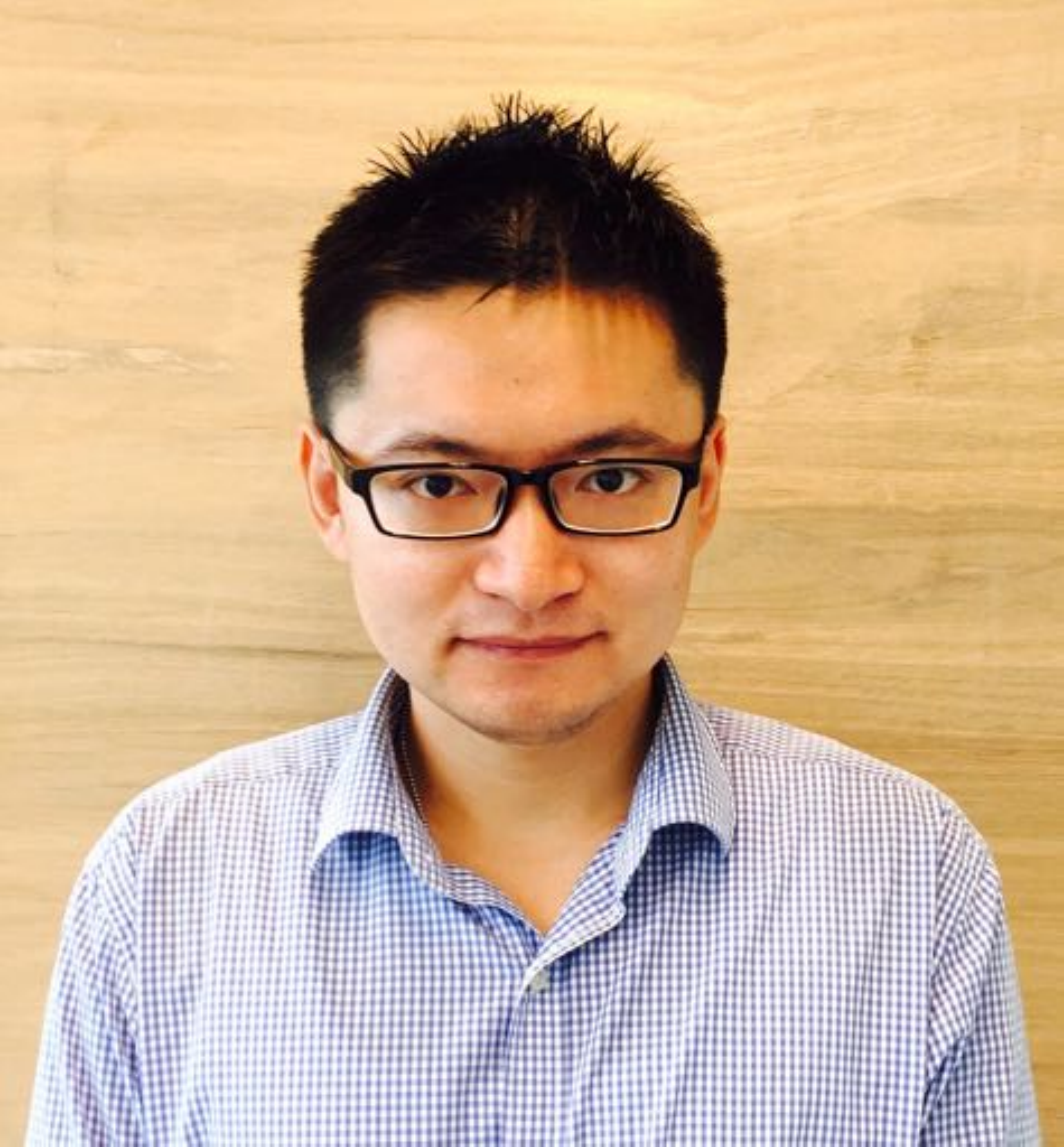}}]{Kai Li} (Senior Member, IEEE) received the B.E. degree from Shandong University, China, in 2009, the M.S. degree from The Hong Kong
University of Science and Technology, Hong Kong, in 2010, and the Ph.D. degree in computer science from The University of New South Wales, Sydney, NSW, Australia, in 2014.

He is currently a Senior Research Scientist with
CISTER, Porto, Portugal. He is a CMU-Portugal Research Fellow, which is jointly supported by Carnegie Mellon University, Pittsburgh, PA, USA, and FCT, Lisbon, Portugal. Prior to this, he was a Postdoctoral Research Fellow with the SUTD-MIT International Design Centre, The Singapore University of Technology and Design, Singapore, from 2014 to 2016. He was a Visiting Research Assistant with ICT Centre, CSIRO, Sydney,
from 2012 to 2013. From 2010 to 2011, he was a Research Assistant with Mobile Technologies Centre, The Chinese University of Hong Kong, Hong Kong. His research interests include machine learning, vehicular communications and security, resource allocation optimization, cyber–physical systems, Internet of Things, and UAV networks.
\end{IEEEbiography}

\begin{IEEEbiography}[{\includegraphics[width=1in,height=1.25in,clip,keepaspectratio]{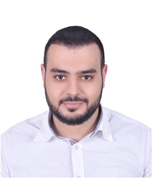}}]{Naram Mhaisen}
received a B.Sc. degree in Computer Engineering, with excellence, from Qatar University (QU) in 2017, and an M.Sc. in Computing from QU in 2020. He then worked as a research assistant at the department of computer science and engineering, QU, and is currently pursuing a Ph.D. at TU Delft. His research interests include the design and optimization of (learning) systems with applications to networking.\end{IEEEbiography}

\begin{IEEEbiography}[{\includegraphics[width=1in,height=1.25in,clip,keepaspectratio]{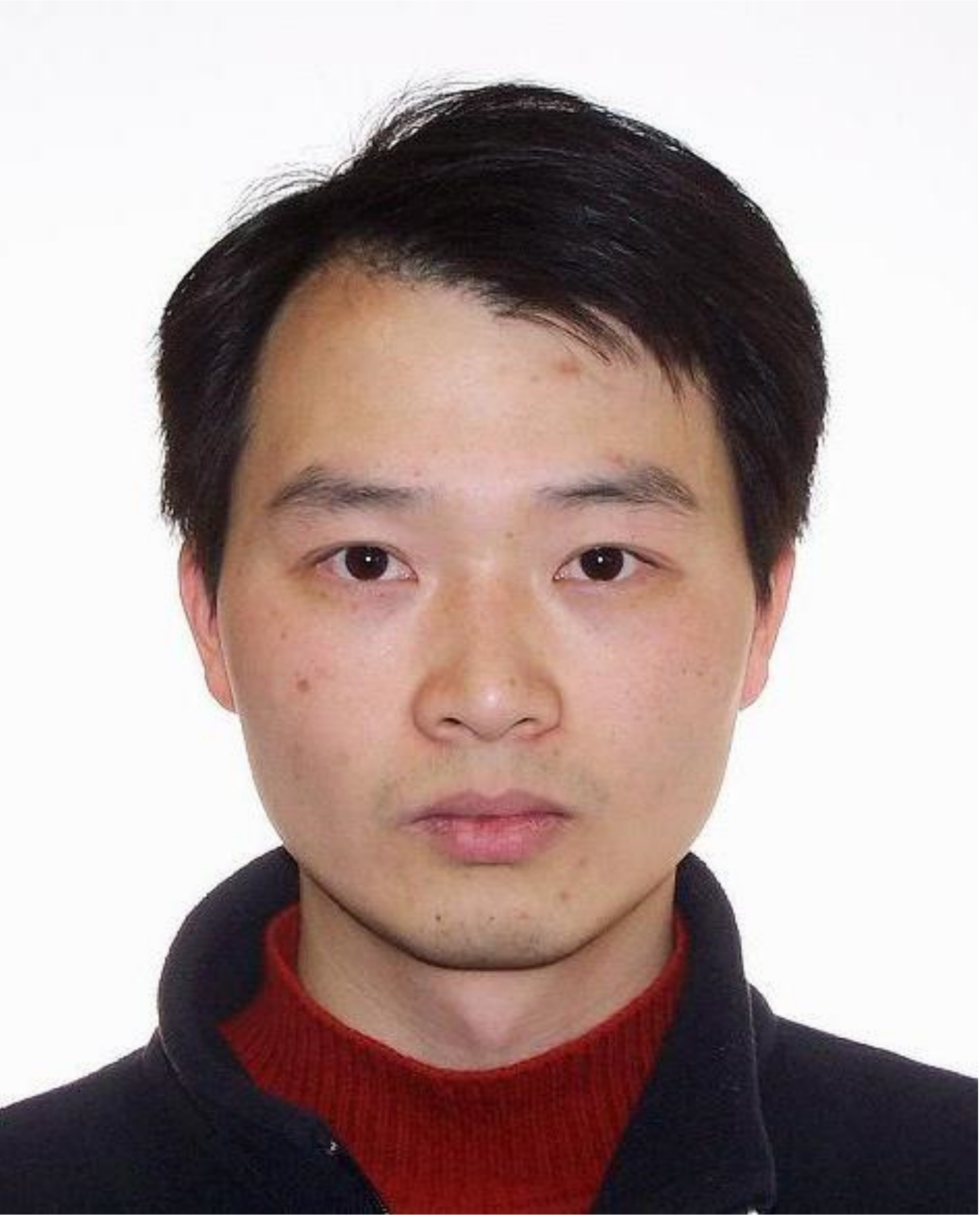}}]{Wei Ni} (Senior Member, IEEE) received the B.E. and Ph.D. degrees in electronic engineering from Fudan University, Shanghai, China, in 2000 and
2005, respectively.

He is currently a Group Leader and a Principal Research Scientist with Commonwealth Scientific and Industrial Research Organization, Sydney, NSW, Australia, an Adjunct Professor with the University of Technology Sydney, Ultimo, NSW, Australia, and an Honorary Professor with Macquarie University, Sydney. Prior to this, he was a Postdoctoral Research Fellow with Shanghai Jiao Tong University, Shanghai, from 2005 to 2008,
and a Deputy Project Manager with Bell Labs, Holmdel, NJ, USA, and Alcatel/Alcatel-Lucent, Boulogne-Billancourt, France, from 2005 to 2008. He was a Senior Researcher with Devices Research and Development, Nokia, Espoo, Finland, from 2008 to 2009. His research interests include signal processing, stochastic optimization, as well as their applications to network
efficiency and integrity. \end{IEEEbiography}

\begin{IEEEbiography}[{\includegraphics[width=1in,height=1.25in,clip,keepaspectratio]{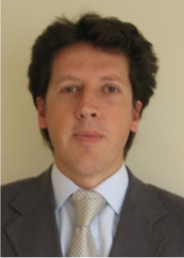}}]{Eduardo Tovar} (Member, IEEE) was born in 1967. He received the Licentiate, M.Sc. and Ph.D. degrees in electrical and computer engineering from the University of Porto, Porto, Portugal, in 1990, 1995, and 1999, respectively. He is currently a Professor with the Computer Engineering Department, the School of Engineering (ISEP) of Polytechnic Institute of Porto (IPP), where he is also engaged in research on real-time distributed systems, wireless sensor networks, multiprocessor systems, cyber-physical systems, and industrial communication systems. He heads the CISTER Research Unit, an internationally renowned research centre focusing on RTD in real-time and embedded computing systems. Since 1991, he authored or coauthored more than 150 scientific and technical papers in the area of real-time and embedded computing systems, with emphasis on multiprocessor systems and distributed embedded systems.  He is currently the Vice-Chair of ACM SIGBED (ACM Special Interest Group on Embedded Computing Systems) and was for five years, until December 2015, member of the Executive Committee of the IEEE Technical Committee on Real-Time Systems (TC-RTS). \end{IEEEbiography}

\begin{IEEEbiography}[{\includegraphics[width=1in,height=1.25in,clip,keepaspectratio]{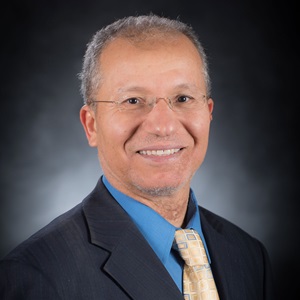}}]{Mohsen Guizani} (M’89–SM’99–F’09) received the BS (with distinction), MS and PhD degrees in Electrical and Computer engineering from Syracuse University, Syracuse, NY, USA in 1985, 1987 and 1990, respectively. He is currently a Professor of Machine Learning and the Associate Provost at Mohamed Bin Zayed University of Artificial Intelligence (MBZUAI), Abu Dhabi, UAE. Previously, he worked in different institutions in the USA. His research interests include applied machine learning and artificial intelligence, Internet of Things (IoT), intelligent systems, smart city, and cybersecurity. He was elevated to IEEE Fellow in 2009 and was listed as a Clarivate Analytics Highly Cited Researcher in Computer Science in 2019, 2020 and 2021. Dr. Guizani has won several research awards including the “2015 IEEE Communications Society Best Survey Paper Award”, the Best ComSoc Journal Paper Award in 2021 as well five Best Paper Awards from ICC and Globecom Conferences. He is the author of ten books and more than 800 publications. He is also the recipient of the 2017 IEEE Communications Society Wireless Technical Committee (WTC) Recognition Award, the 2018 AdHoc Technical Committee Recognition Award, and the 2019 IEEE Communications and Information Security Technical Recognition (CISTC) Award. He served as the Editor-in-Chief of IEEE Network and is currently serving on the Editorial Boards of many IEEE Transactions and Magazines. He was the Chair of the IEEE Communications Society Wireless Technical Committee and the Chair of the TAOS Technical Committee. He served as the IEEE Computer Society Distinguished Speaker and is currently the IEEE ComSoc Distinguished Lecturer.\end{IEEEbiography}

\end{document}